\documentclass{article}

\usepackage{microtype}
\usepackage{graphicx}
\usepackage{subcaption}
\usepackage{booktabs}
\usepackage{tabularx}
\usepackage{multirow}
\usepackage{colortbl}
\usepackage{pifont}
\usepackage[most]{tcolorbox}
\usepackage{listings}

\usepackage{amsmath}
\usepackage{amssymb}
\usepackage{mathtools}
\usepackage{amsthm}

\usepackage[accepted]{arxiv/preprint}

\usepackage{hyperref}
\usepackage[capitalize,noabbrev]{cleveref}

\theoremstyle{plain}
\newtheorem{theorem}{Theorem}[section]

\theoremstyle{definition}

\newtheorem{remark}[theorem]{Remark}

\usepackage{amsmath,amsfonts,bm}

\newcommand{\EE}{\mathbb{E}}
\newcommand{\RR}{\mathbb{R}}

\input{commands}

\newcommand{\ra}[1]{\renewcommand{\arraystretch}{#1}}
\lstdefinestyle{prompttext}{
  basicstyle=\ttfamily\fontsize{6.5}{6.5}\selectfont,
  breaklines=true,
  breakatwhitespace=false,
  columns=fullflexible,
  keepspaces=true,
  showstringspaces=false,
  aboveskip=0pt,
  belowskip=0pt
}
\tcbset{
  promptbox/.style={
    enhanced,
    breakable,
    colback=gray!3,
    colframe=black!45,
    boxrule=0.4pt,
    arc=1pt,
    left=4pt,
    right=4pt,
    top=2pt,
    bottom=2pt,
    width=0.95\linewidth
  }
}

\hypersetup{pdfkeywords={Video Editing, Diffusion Models, Layered Generation}}

\runningtitle{Vera: A Layered Diffusion Model for Content-Preserving Video Editing}

\title{Vera: A Layered Diffusion Model for Content-Preserving Video Editing}

\author{
  Hongkai Zheng\textsuperscript{1,2\,\textdagger}\quad
  Ta-Ying Cheng\textsuperscript{2}\quad
  Benjamin Klein\textsuperscript{2}\\[2pt]
  Yisong Yue\textsuperscript{1}\quad
  Zhuoning Yuan\textsuperscript{2\,\textdaggerdbl}\\[6pt]
  {\normalsize
    \textsuperscript{1}California Institute of Technology\quad
    \textsuperscript{2}Netflix, Inc.}
}

\date{}

\begin{document}

\maketitle

\vspace{-0.3in}
\begin{center}
\href{https://vera-layered-diffusion.github.io/}{\textcolor[HTML]{D62B70}{https://vera-layered-diffusion.github.io/}}   
\end{center}
\vspace{0.2in}

\begingroup
\renewcommand\thefootnote{}%
\NoHyper
\footnotetext{%
  \begin{tabular}[t]{@{}r@{\ }l@{}}
    \textdagger     & Work done during an internship at Netflix.\\
    \textdaggerdbl  & Project Lead. Correspondence: Zhuoning Yuan \texttt{<zyuan@netflix.com>}.
  \end{tabular}}%
\endNoHyper
\endgroup

\begin{abstract}
Video diffusion models have enabled remarkable progress in video generation and editing. However, content preservation remains a core challenge: existing methods regenerate every pixel and often alter elements that should remain unchanged, such as characters or background scenes.
We introduce Vera, a layered diffusion framework for content-preserving video editing. Instead of regenerating the entire video, Vera generates an edit layer along with an alpha matte for compositing with the source video, separating creative editing from content preservation by design. To encourage coherent composition with the source video, we extend the text-to-video DiT into a Mixture-of-Transformers (MoT) architecture, with separate DiTs for each layer that interact through joint self-attention.
To support the training of Vera, we further construct a high-quality layered dataset with accurate alpha mattes, diverse scenes and dynamics, and visual effects. 
Across our quantitative benchmark and human preference study, Vera outperforms leading open-source video editing models in content preservation while remaining competitive in edit quality, using 486K frames of layered training data.
\end{abstract}

\section{Introduction}
\begin{figure}[ht!]
  \centering
  \includegraphics[width=0.96\linewidth]{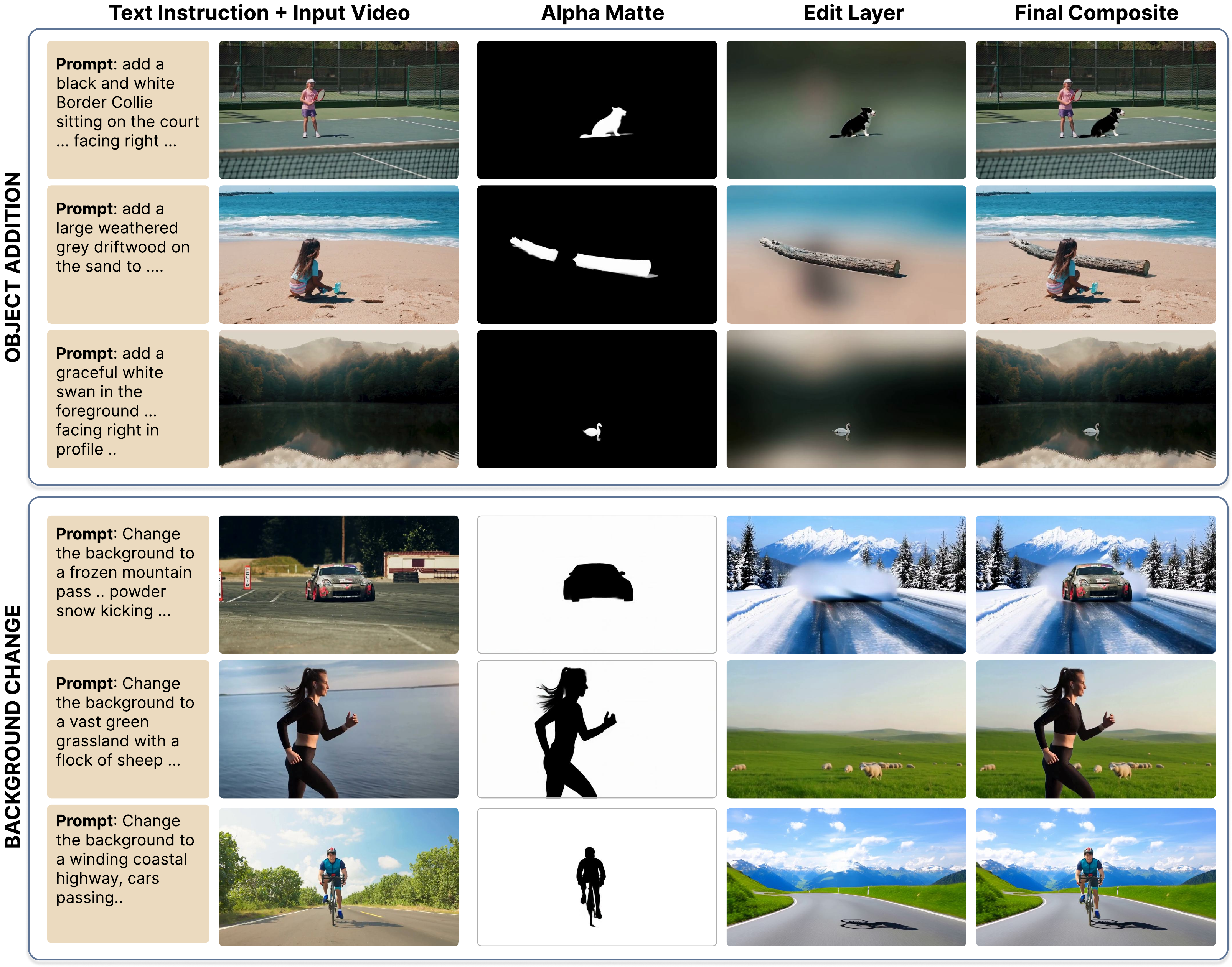}
  \caption{Given an input video and a text instruction, Vera generates an edit layer together with an alpha matte that can be directly composited with the input video to produce the edited result. For object addition, the alpha includes the effects (\eg shadows) to be added into the composite; for background replacement, Vera learns to include the effects that complements the preserved regions (\eg smoke behind cars). The joint generation of edit layer and its matte allows the original content to be preserved during editing, even for very fine-grained details (\eg hair of the girl during background change). The input prompts in the figure have been shortened for visual clarity.}
  \label{fig:results_bg_bike}
\end{figure}

Video generation has recently achieved remarkable breakthroughs with cinematic-level quality~\cite{veo3_2026,openai2025sora2}. Driven by advancements in text-to-video (T2V) foundation diffusion models~\cite{wan2025wan, cogvideox_2024,kong2024hunyuanvideo,HaCohen2024LTXVideo}, controllable video editing has attracted significant attention for narrowing the gap between video generation and practical production use. Recent progress spans object insertion and removal, background replacement, visual effects (VFX), style transfer, and relighting~\cite{bian2025videopainter, tang2023any, zhang2025reco, litman2026editctrl, fu2025layeredit, li2025vfxmaster, zhou2025light}. These capabilities streamline complex manual editing tasks that traditionally take weeks, and lower the barrier for creators to focus on creativity rather than technical execution. Despite the progress, a key challenge remains  \textbf{content-preserving editing}: regions outside the target edited area are often inadvertently modified. In commercial production, even a small unintended change can render an edit unusable and erode user trust in the tool.

Existing approaches attempt to address this through various strategies such as regional constraints and mask conditioning~\cite{zhang2025reco, jiang2025vace, litman2026editctrl, jia2025cococo}, aiming to isolate targeted edits from unrelated regions. 
However, these methods still operate within the end-to-end diffusion paradigm, which has two fundamental limitations: (i) even with explicit region control, the model regenerates the entire video and can still introduce inadvertent changes to regions that should be strictly preserved, especially in complex scenarios; (ii) the end-to-end paradigm produces only a final composite, whereas practical production workflows often require layered assets for iterative editing—manually separating added content from the output for subsequent edits is tedious and error-prone. 

Layered approach~\cite{yin2025qwen, ji2025layerflow, wang2025transpixeler, dong2025wanalpha} offers a principled alternative by producing edit content as a separate layer with an alpha matte for compositing, preserving the original video by construction. However, existing work in this direction has focused primarily on text-to-RGBA video generation rather than editing, and applying layered generation to video editing introduces significant new challenges and largely remains underexplored. The edit layer and alpha matte must be precisely aligned, and the generated content must be highly consistent with the source video—matching camera motion, lighting, spatial layout, and scale—to achieve natural compositing. The model must also handle complex cross-layer interactions such as shadows, reflections, and occlusions. These challenges demand both an architecture capable of coherent cross-layer generation and high-quality layered training data that captures such interactions.

\begin{figure}[t]
    \centering
    \includegraphics[width=0.8\linewidth]{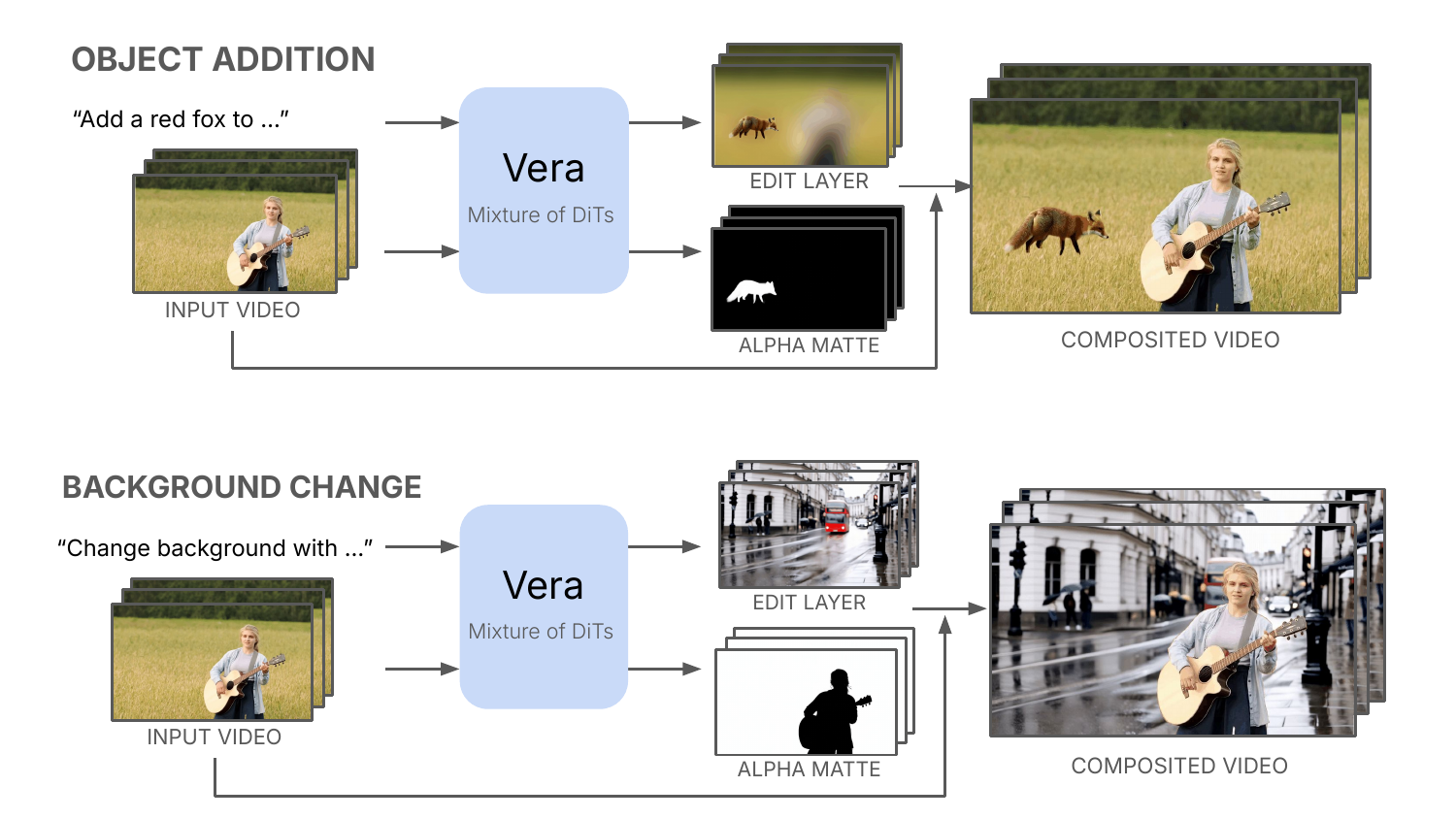}
    \caption{Overview of the Vera inference pipeline. Given an input video and a text editing instruction, Vera's MoT architecture jointly generates an edit layer, an alpha matte, and a composite video. The edit layer and alpha matte are then composited with the source video to produce the final edited output. }
    \label{fig:diagram}
\end{figure}

We introduce \textbf{Vera} (from the Latin vēra, meaning "genuine"), a layered diffusion framework that addresses these challenges for content-preserving video editing.
As depicted in Fig.~\ref{fig:diagram}, given a source video and an editing instruction, Vera generates three outputs: an edit layer, an alpha matte, and a composite video.
The edit layer and alpha matte composite with the source video to produce the final result, explicitly separating the generated edit from the original content.
To encourage coherent composition with the source video, we extend the text-to-video DiT into a MoT architecture with separate DiTs per layer that interact through joint self-attention, and curate a high-quality layered dataset with accurate alpha mattes, diverse dynamics, and visual effects. Our contributions are as follows:

\begin{itemize}
\item We propose Vera, a new layered video editing framework
that preserves content integrity by generating edits as a separate RGBA layer, enabling both faithful preservation and natural composition. 
\item We construct a high-quality layered video dataset comprising synthetic composites, realistic single- and multi-object scenes, and scenes with interactive visual effects, along with a test benchmark spanning diverse motion dynamics and scene complexity.
\item Across our quantitative benchmark and human preference study, Vera outperforms leading open-source video editing models in content preservation while remaining competitive in edit quality, using 486K frames of layered training data.
\item Through controlled experiments, we identify the key architecture and data choices that enable a strong layered diffusion model with both faithful preservation and high-quality composition.
\end{itemize}

\section{Related Work}

\subsection{Diffusion Models for Video Generation and Editing}
The advent of diffusion models~\cite{ho2020ddpm,liu2023rectified} transformed the landscape of generative vision models, revolutionizing high-fidelity image synthesis~\cite{sd3_2024, flux_2025} and soon video generation~\cite{veo3_2026, runway_gen_2023, wan2025wan}. Recently, proprietary models such as Runway's Gen series~\cite{runway_gen_2023, runway_gen4_2025} and Google's Veo 3~\cite{veo3_2026} have shown powerful capabilities, offering highly controllable text-to-video generation and editing. Concurrently, several open-source video models such as LTX-Video~\cite{HaCohen2024LTXVideo}, CogVideo~\cite{cogvideox_2024}, and WAN~\cite{wan2025wan}, have also shown promising capabilities in generating temporally consistent videos.

Adapting these powerful generative priors for video editing has also become a highly active area of research. Several approaches now enable temporally consistent, prompt-driven video editing~\cite{jiang2025vace, geyertokenflow}. Furthermore, the scope of editing has expanded toward end-to-end visual effects (VFX) generation~\cite{li2025vfxmaster, bian2025video}, allowing for complex structural and stylistic scene modifications. To support the training and evaluation of these complex editing frameworks, comprehensive video editing datasets have also been recently introduced to the community~\cite{zisenorita, hu2024vivid, zhang2025reco}. However, one major drawback of these models is the faithfulness of unchanged regions – the nature of denoising a new video probabilistics leads to many unintended changes in places not imposed by the text prompt. Vera aims to mitigate this issue by proposing an end-to-end layered framework that generates both the edited video and its corresponding alpha matte.

\subsection{Layer-wise Image Video Generation}
To achieve granular control while maintaining faithfulness over scene composition, some recent work has gravitated toward layer-wise image and video generation strategies. LayerDiffuse~\cite{zhang2024transparent} is one of the first to generate transparent images in multiple transparent layers. LayerFusion \cite{dalva2024layerfusion} generates images with two layers jointly -- a foreground RGBA layer with a background layer -- to enhance the harmonization. LayerDecomp \cite{yang2025generative} decomposes an image into RGB layer containing effects like shadows for edits. LayerEdit~\cite{fu2025layeredit} proposes a training-free method that decomposes objects into layers through the model's attention for more disentangled multi-object editing. More recently, Qwen-Image-Layered~\cite{yin2025qwen} decomposes an image into multiple RGBA layers via a multimodal diffusion transformer.

Other work have extended transparent and layered visual generation into videos. LayerFlow \cite{ji2025layerflow} is one of the first to propose layer-wise video generation from per-layer prompts. Transpixeler \cite{wang2025transpixeler} builds on top of pretrained video models into additional outputing alpha channel while keeping its original RGB capabilities. Wan-Alpha \cite{dong2025wanalpha} learns to shift away the alpha distribution from the RGB distribution to enable better RGBA video generation. On the other hand, notable works like Generative Omnimatte \cite{lee2025generative} decomposes videos into multiple layers, which can then be used for video editing. These methods, however, lack the high-quality video data for training and struggle with complex interactions like shadows and reflections.

\section{Method}
\begin{figure}[t]
  \centering
  \includegraphics[width=0.95\linewidth]{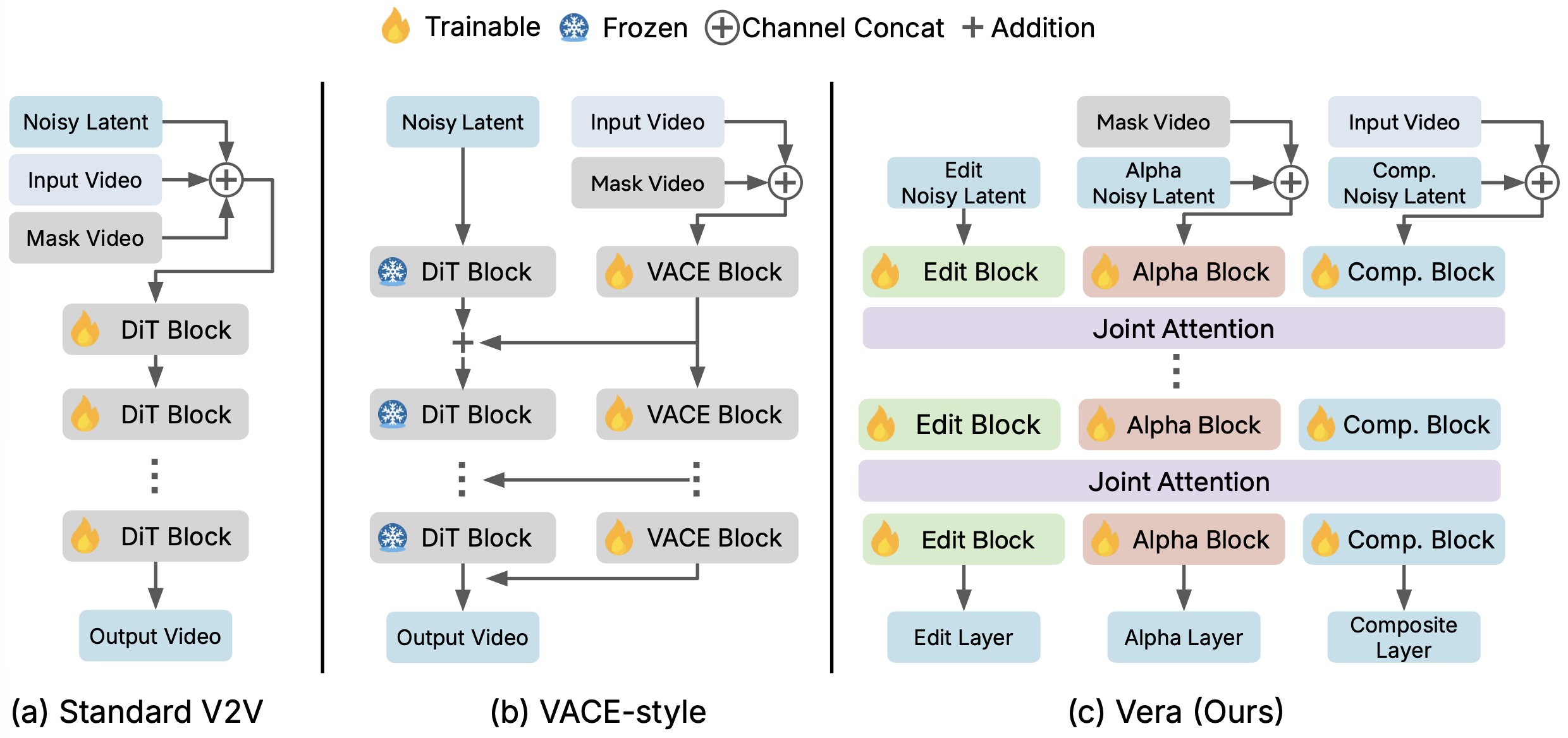}
  \caption{The architecture of Vera compared to other video editing methods. VAE encoding, VAE decoding, and patchifying are omitted for clarity. (a) Standard fine-tuning of a pretrained T2V model for video editing. (b) VACE-style~\cite{jiang2025vace} fine-tuning with additional context adapter blocks. (c) Vera consists of three DiTs, each responsible for modeling a separate layer, with interactions across layers enabled through joint self-attention. Ablations on this is presented in \cref{sec:ablation}.}
  \label{fig:architecture}
\end{figure}

We consider the content-preserving video editing problem, where the source video $V_{\mathrm{src}} \in \RR^{T\times H\times W\times 3}$ is a composition of two parts: 
\begin{equation}\label{eq:decomposition}
  V_{\mathrm{src}} = (1-A_{\mathrm{edit}}) \circ V_{\mathrm{preserved}} + A_{\mathrm{edit}} \circ V_{\mathrm{non-preserved}}, 
\end{equation}
where $V_{\mathrm{preserved}}\in \RR^{T\times H\times W\times 3}$ is the content that should remain untouched, $A_{\mathrm{edit}}\in [0,1]^{T\times H\times W}$ is an alpha matte, and $\circ$ denotes the Hadamard product. 
Given the source video $V_{\mathrm{src}}$ and a conditioning signal $C$ (\eg, a text prompt and an optional mask), our goal is to generate an edit layer $V_{\mathrm{edit}}$ along with the corresponding alpha matte $A_{\mathrm{edit}}$. The output video is then composited as: 
\begin{equation}
  \label{eq:problem}
  V_{\mathrm{composite}} =  (1- A_{\mathrm{edit}}) \circ V_{\mathrm{preserved}} + A_{\mathrm{edit}} \circ V_{\mathrm{edit}} .
\end{equation}
While $A_{\mathrm{edit}}$ and $V_{\mathrm{preserved}}$ are largely constrained by the source video and conditioning signal, they are not directly available and must be inferred by the model. In binary-mask regions, $V_{\mathrm{preserved}}$ is fully determined by the mask and the source video. In semi-transparent regions, however, the source video is a mixture of preserved and non-preserved content, and the model must disentangle them. In this work, we focus on the common setting where semi-transparent regions are small so that $V_{\mathrm{preserved}}$ can be well approximated by $V_{\mathrm{src}}$. This layered formulation explicitly separates creative edit generation from content preservation, maintaining the integrity of the original video.

\begin{remark}\label{remark:semi-transparent}
The above approximation requires that $V_{\mathrm{preserved}}$ has minimal semi-transparent regions, which holds for common editing applications, including background replacement and object addition. Importantly, the edit layer $V_{\mathrm{edit}}$ is fully generative and may contain large semi-transparent regions (\eg, shadows, reflections, or glasses) without affecting the approximation. Our three-output formulation can in principle represent cases where $V_{\mathrm{preserved}}$ contains substantial semi-transparent regions. Demonstrating this capability, however, requires suitable layered training data, and we do not evaluate this setting in this work.
\end{remark}

\subsection{Vera Framework and Architecture}

\begin{figure}[ht!]
  \centering
  \includegraphics[width=0.99\linewidth]{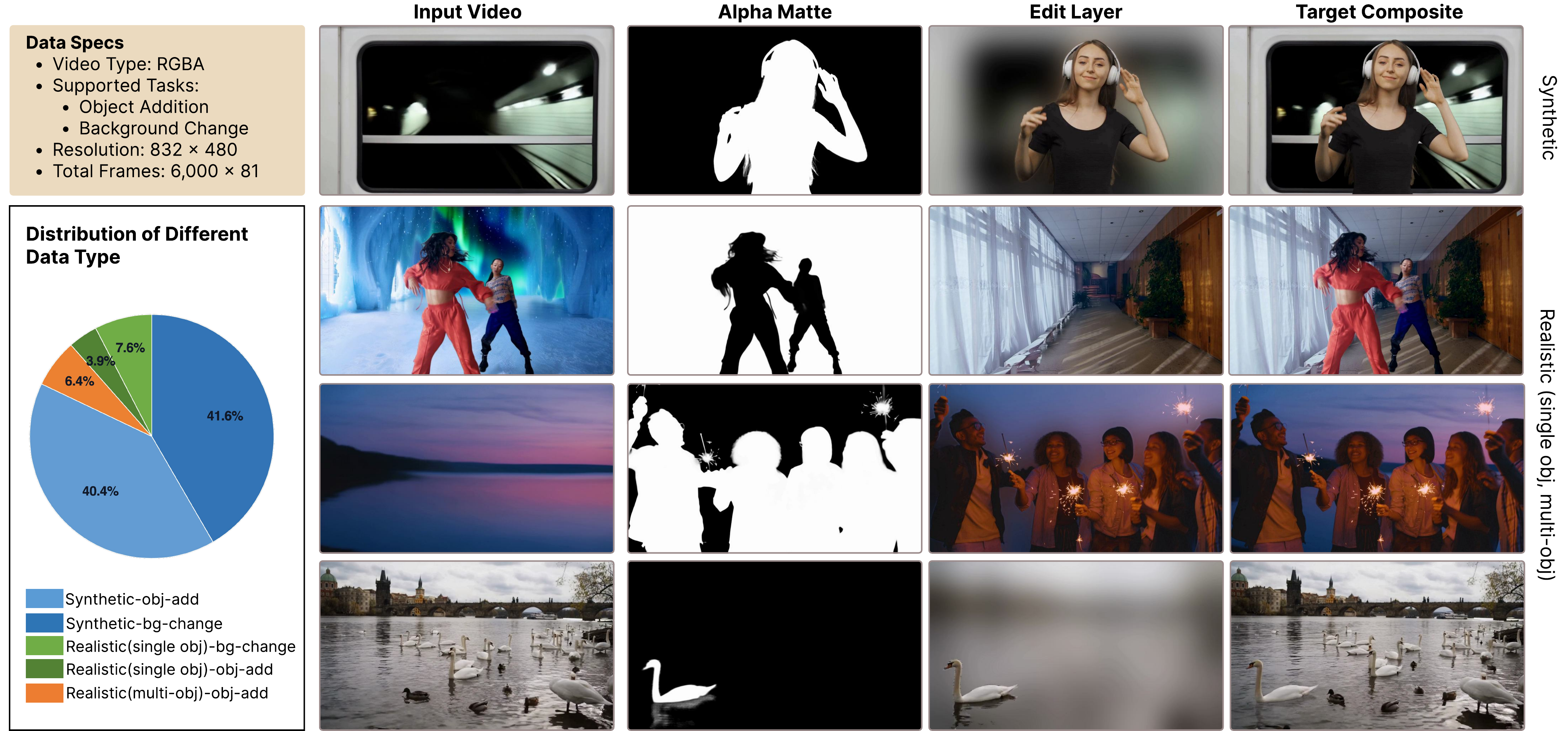}
  \caption{Overview of our layered training data. Each sample consists of an input video, an edit layer, an alpha matte, and a composite target video. The white regions in the edit layer indicate transparency. We curate data for two tasks: background change and object addition, including samples with interactive effects such as shadows and reflections.}
  \label{fig:data-preview}
  
  \vspace{1cm} %
  
    \includegraphics[width=0.99\linewidth]{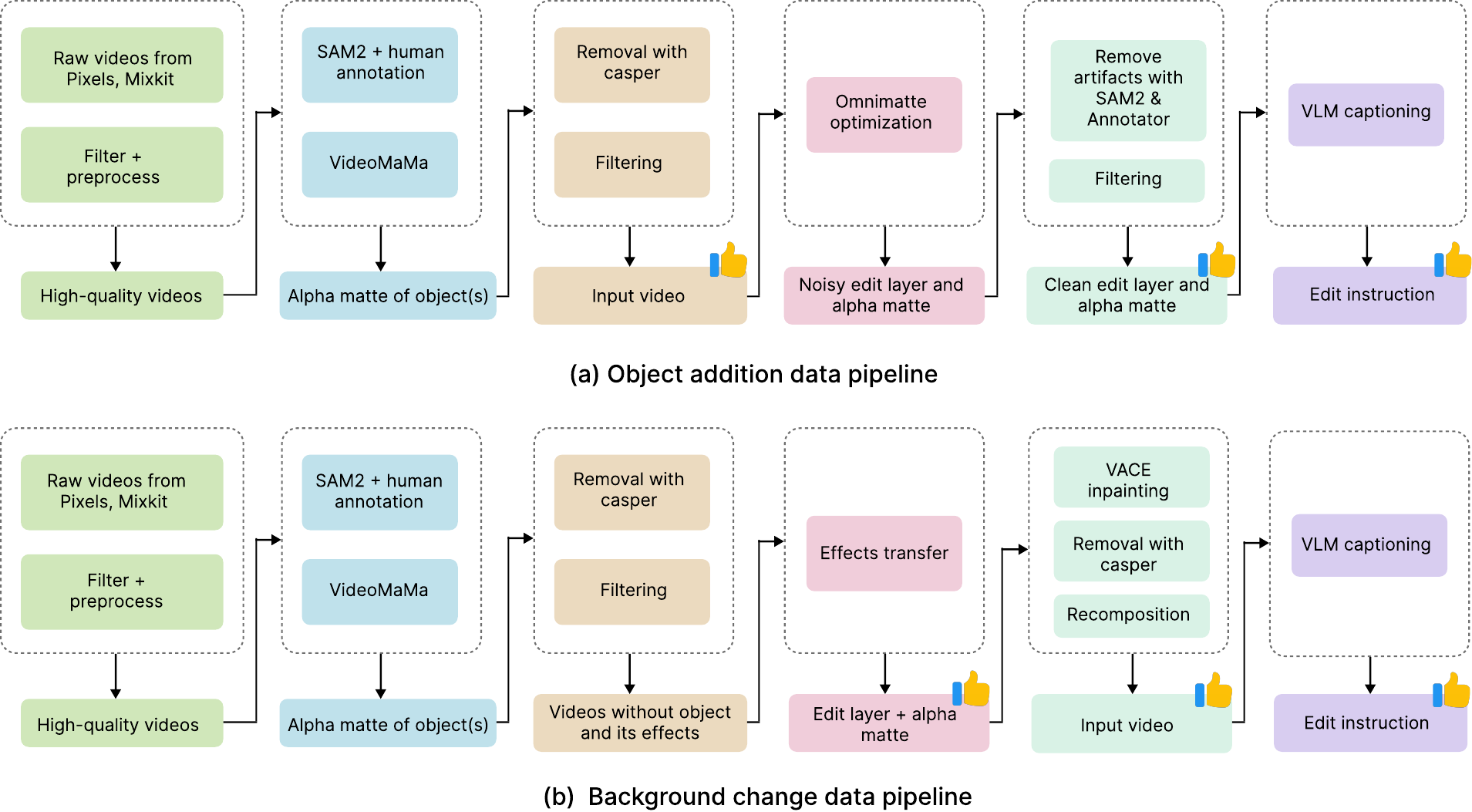}
\caption{Overview of the data construction pipelines for (a) object addition and (b) background change. Each color represents a distinct stage; dashed-line blocks denote the operations within a stage, and the block outside the dashed line is the stage output. Thumb-up icons denote the final outputs to be used for model training and evaluations.}
\label{fig:data-pipelines}
\end{figure}

\subsubsection{Modeling framework}
By the compositing equation (Eq.~\ref{eq:problem}), any three of the four variables $V_{\mathrm{edit}}$, $A_{\mathrm{edit}}$, $V_{\mathrm{preserved}}$, and $V_{\mathrm{composite}}$ fully determine the fourth. We therefore model the joint conditional distribution $p(V_{\mathrm{edit}}, A_{\mathrm{edit}}, V_{\mathrm{composite}} \mid V_{\mathrm{src}}, C)$ using a diffusion model~\cite{songscore}. We choose to include $V_{\mathrm{composite}}$ over $V_{\mathrm{preserved}}$ because $V_{\mathrm{composite}}$ shares the distribution of natural videos, which aligns better with the pretrained video generation model. Once these three quantities are generated, the preserved content can be recovered as:
\begin{equation}
  \label{eq:recover}
  V_{\mathrm{preserved}} = 
  \begin{cases}
    \dfrac{V_{\mathrm{composite}} - A_{\mathrm{edit}} \circ V_{\mathrm{edit}}}{1 - A_{\mathrm{edit}}}, & \text{in semi-transparent regions}, \\[6pt]
    V_{\mathrm{src}}, & \text{elsewhere}.
  \end{cases}
\end{equation}
In our experiments, we use the approximation $V_{\mathrm{preserved}} \approx V_{\mathrm{src}}$. Note that $V_{\mathrm{edit}}$ is mathematically undefined in the region where $A_{\mathrm{edit}}=0$. We set the undefined region to a Gaussian smoothed version of the composite for better edge blending and consistency.

We implement this with a latent diffusion model framework. 
All videos are encoded into a latent space using the Wan2.1 VAE~\cite{wan2025wan}, where the alpha matte sequence is treated as a video with identical RGB channels. We denote the latents of $V_{\mathrm{src}}$, $V_{\mathrm{edit}}$, $A_{\mathrm{edit}}$, and $V_{\mathrm{composite}}$ as $Z_{\mathrm{src}}$, $Z_{\mathrm{edit}}$, $Z_{\mathrm{alpha}}$, and $Z_{\mathrm{composite}}$, respectively. Using the flow matching~\cite{lipmanflow, liu2023rectified} parameterization, we train a neural network $u_\theta$ that takes $Z_{\mathrm{src}}$ and $C$ as inputs and jointly predicts the velocity fields for all the layers: $u_\theta=[u_{\theta; \mathrm{edit}},u_{\theta; \mathrm{alpha}},u_{\theta; \mathrm{composite}}]$.
The training objective is:  
\begin{equation}
  \mathcal{L}_{\mathrm{FM}} = \EE_{t, C, Z_\mathrm{src}, Z_{1}, Z_{0}} \|u_{\theta}(Z_t;t, C, Z_{\mathrm{src}}) - (Z_{1} - Z_{0}) \|_2^2,
\end{equation}
where $Z_{0} \sim \mathcal{N}(0, I)$, $Z_1 = [Z_{\mathrm{edit}}, Z_{\mathrm{alpha}}, Z_{\mathrm{composite}}]$, $t\in[0, 1]$, and $Z_t = t Z_1 + (1-t) Z_0$.

\subsubsection{Architecture design}
A key design challenge is that the three generated quantities (edit layer $V_{\mathrm{edit}}$, alpha matte $A_{\mathrm{edit}}$, and composite video $V_{\mathrm{composite}}$) have substantially different distributions: $V_{\mathrm{edit}}$ contains decoupled creative content, $A_{\mathrm{edit}}$ is a grayscale matte that depends not only on the edit content but also on interactions between the edit and the original scene (\eg, an inserted object partially occluded by a foreground subject), and $V_{\mathrm{composite}}$ is a natural video. A single shared transformer would need to reconcile these distributional differences entirely through training, which we found to be data-inefficient~\cite{team2024chameleon}.

Following the Mixture-of-Transformers (MoT) framework~\cite{liang2025mixtureoftransformers}, we use three separate DiTs—one per output—that interact through joint self-attention as shown in Fig.~\ref{fig:architecture}. While MoT was originally proposed for multi-modal learning, we find it equally effective for modeling the interactions between output layers with distinct distributions. Each DiT maintains its own QKV projections and FFN weights, but tokens from all three DiTs are concatenated into a single sequence for the self-attention operation, enabling cross-layer interaction while allowing each branch to specialize.

All three DiTs are initialized from the same pretrained text-to-video model~\cite{wan2025wan}. To incorporate conditioning video inputs, we introduce two additional patch embedding layers: one for the input video and one for an optional mask video. The source video tokens are added to the composite tokens, while the mask tokens are added to the noisy alpha tokens. All layer tokens share the same positional encoding (RoPE). To allow the model to distinguish between layers, we add a zero-initialized learnable embedding to the alpha and composite layer tokens, respectively.

\subsection{Data Construction and Curation}
Since no public dataset provides layered video editing data suitable for training our model, we construct a layered dataset from open-source videos using a combination of annotation and generation tools. 
Fig.~\ref{fig:data-preview} shows an overview of our layered training data. The entire dataset contains 486K frames in $832\times 480$ resolution, categorized into complementary subsets of increasing complexity (see pie chart in Fig.~\ref{fig:data-preview}).
It comprises roughly 6K samples, each a tuple of four videos---one input plus three output layers (edit, alpha, and composite)---at $832\times480$ and 81 frames; about 60--70\% of filtered source videos yielded usable samples through our pipeline.
We give a brief overview of each subset below; the data pipelines are illustrated in Fig.~\ref{fig:data-pipelines} and described in detail in Appendix~\ref{sec:appendix-data-pipeline}.

\noindent\textbf{Synthetic composites (Object Addition and Background Change).}
We derive layered foreground-background data from VideoMatte240K~\cite{lin2021real}, which provides carefully annotated alpha mattes with precise alpha values for fine structures such as hair -- details that are difficult to obtain from automated matting tools. 
Since VideoMatte240K contains only foreground mattes without backgrounds, we generate diverse synthetic backgrounds using an inpainting model~\cite{jiang2025vace} and composite them with the extracted foregrounds. This subset provides accurate alpha supervision for both background change and object addition, but is limited to human subjects filmed with centered, static cameras.

\noindent\textbf{Realistic single-object videos (Object Addition and Background Change).}
To introduce natural scene diversity and camera motion, we source real-world videos from Pexels~\cite{pexels} and Mixkit~\cite{mixkit}. We build a multi-stage data pipeline that generates layered data via a chain of segmentation~\cite{ravisam}, video matting~\cite{lim2026videomama}, video inpainting and generation~\cite{jiang2025vace,zi2025minimax}, and human annotation and filtering. These videos feature diverse scenes and dynamic camera motion, but typically contain a single prominent subject with limited visual effects. This subset provides training data for both background change and object addition.

\noindent\textbf{Realistic multi-object videos with effects (Object Addition only).}
To handle more complex scenes, we extend the pipeline above with an additional omnimatte optimization step~\cite{lee2025generative} and an alpha matte curation stage, enabling the extraction of individual objects together with their associated visual effects (\eg, shadows and reflections). This subset features multiple objects in complex scenes, rich dynamics, and provides data exclusively for the object addition task.

\section{Experiments}
\begin{table}[t]
  \centering
  \caption{Comparison with existing video editing methods. VLM-based metrics (CS, CT, IS) are averaged over three VLMs. OC and TF are near-identical across many models and not bolded. Best results in other columns are in \textbf{bold}. 
  LayerFlow is included for reference only, as it is limited to 16 frames at 720$\times$480 resolution and thus evaluated under different conditions.}
  \label{tab:main-comparison}
  \small
  \ra{1.1}
  \resizebox{\linewidth}{!}{
  \begin{tabular}{lccccccccc}
  \toprule
  \multirow{2}{*}{Method}
    & \multicolumn{3}{c}{\shortstack{Content\\Preservation}}
    & \multicolumn{4}{c}{\shortstack{Video\\Quality}}
    & \multicolumn{2}{c}{\shortstack{Instruction\\Compliance}} \\
  \cmidrule(lr){2-4} \cmidrule(lr){5-8} \cmidrule(lr){9-10}
    & PSNR $\uparrow$ & SSIM $\uparrow$ & LPIPS $\downarrow$
    & OC $\uparrow$ & TF $\uparrow$ & CS $\uparrow$ & CT $\uparrow$
    & OSC $\uparrow$ & IS $\uparrow$ \\
  \midrule
  \multicolumn{10}{l}{\textit{Object Addition}} \\
  \midrule
  Ditto~\cite{bai2025scaling} & 13.1 & 0.447 & 0.529 & 0.98 & 0.98 & 2.84 & 3.13 & 0.239 & 3.62 \\
  Lucy-Edit~\cite{decart2025lucyedit} & 19.0 & 0.798 & 0.301 & 0.98 & 0.98 & 2.54 & 2.79 & 0.238 & 2.98 \\
  ICVE~\cite{liao2025context} & 18.0 & 0.818 & 0.265 & 0.98 & 0.98 & 2.88 & 3.12 & 0.239 & 3.31 \\
  VideoPainter~\cite{bian2025videopainter} & 10.7 & 0.354 & 0.669 & 0.64 & 0.98 & 2.54 & 2.80 & 0.247 & 2.87 \\
  VACE-1.3B~\cite{jiang2025vace} & 10.3 & 0.323 & 0.667 & 0.98 & 0.98 & 2.55 & 3.01 & \textbf{0.250} & 2.99 \\
  VACE-14B~\cite{jiang2025vace} & 10.1 & 0.325 & 0.666 & 0.98 & 0.98 & 2.63 & 3.08 & 0.248 & 3.26 \\
  ReCo~\cite{zhang2025reco} & 16.8 & 0.778 & 0.280 & 0.98 & 0.98 & 3.18 & 3.35 & 0.243 & 3.77 \\
  \rowcolor{blue!7}
  Vera-1.3B & 25.3 & 0.949 & \textbf{0.078} & 0.98 & 0.98 & 3.49 & 3.47 & 0.243 & 4.05 \\
  \rowcolor{blue!7}
  Vera-14B & \textbf{26.1} & \textbf{0.950} & 0.082 & 0.98 & 0.98 & \textbf{3.59} & \textbf{3.66} & 0.244 & \textbf{4.13} \\
    \midrule
  \multicolumn{10}{l}{\textit{Background Change}} \\
  \midrule
  Ditto~\cite{bai2025scaling} & 21.7 & 0.888 & 0.092 & 0.97 & 0.97 & 3.29 & 3.25 & 0.240 & 4.26 \\
  Lucy-Edit~\cite{decart2025lucyedit} & 22.3 & 0.908 & 0.078 & 0.96 & 0.96 & 2.50 & 2.50 & 0.235 & 3.71 \\
  ICVE~\cite{liao2025context} & 26.3 & 0.950 & 0.050 & 0.96 & 0.97 & 2.80 & 2.97 & 0.228 & 3.78 \\
  VideoPainter~\cite{bian2025videopainter} & 29.9 & 0.975 & 0.032 & 0.83 & 0.97 & 2.29 & 2.31 & 0.237 & 3.64 \\
  VACE-1.3B~\cite{jiang2025vace} & 31.5 & 0.981 & 0.024 & 0.96 & 0.97 & \textbf{3.74} & \textbf{3.61} & \textbf{0.245} & \textbf{4.48} \\
  VACE-14B~\cite{jiang2025vace} & 31.7 & 0.983 & 0.021 & 0.96 & 0.96 & 3.68 & 3.59 & 0.243 & 4.44 \\
  ReCo~\cite{zhang2025reco} & 24.9 & 0.935 & 0.061 & 0.96 & 0.97 & 2.45 & 2.71 & 0.220 & 3.05 \\
  LayerFlow*~\cite{ji2025layerflow} & 23.3 & 0.886 & 0.087 & 0.94 & 0.95 & 2.51 & 2.35 & 0.243 & 4.14 \\
  \rowcolor{blue!7}
  Vera-1.3B & 35.2 & \textbf{0.993} & \textbf{0.010} & 0.96 & 0.97 & 3.34 & 3.25 & 0.243 & 4.35 \\
  \rowcolor{blue!7}
  Vera-14B & \textbf{36.2} & \textbf{0.993} & \textbf{0.010} & 0.96 & 0.97 & 3.28 & 3.21 & 0.242 & 4.38 \\
  \bottomrule
  \end{tabular}
  }
\end{table}

\begin{figure}[h]
    \centering
    \includegraphics[width=0.98\linewidth]{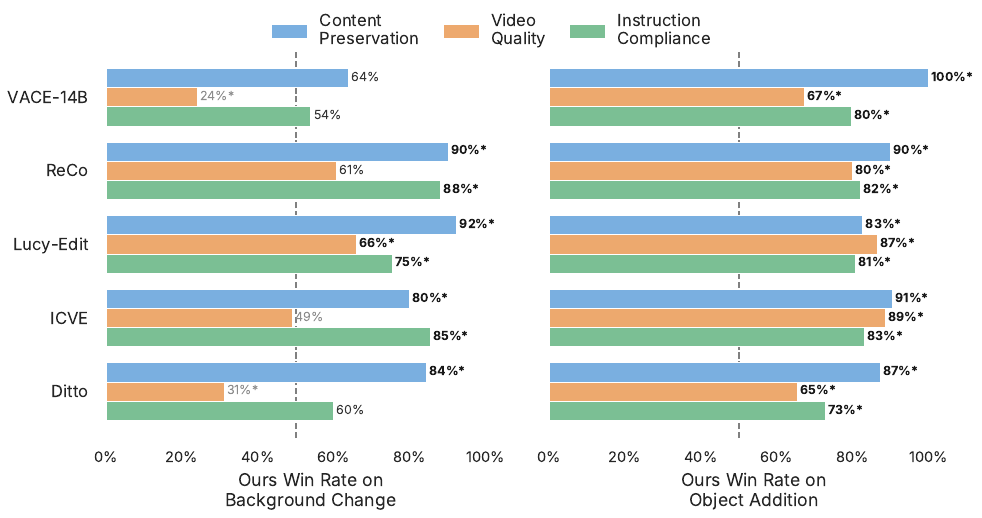}
    \caption{2AFC user study results comparing Vera-1.3B against five baselines. Bars show our win rate across three 
    evaluation dimensions. Bold values with $\*$ indicate statistically significant results (p < 0.05, binomial test) where our model is preferred.}
    \label{fig:user-study-winrates}
\end{figure}

\begin{figure}[h]
  \centering
  \includegraphics[width=0.99\linewidth]{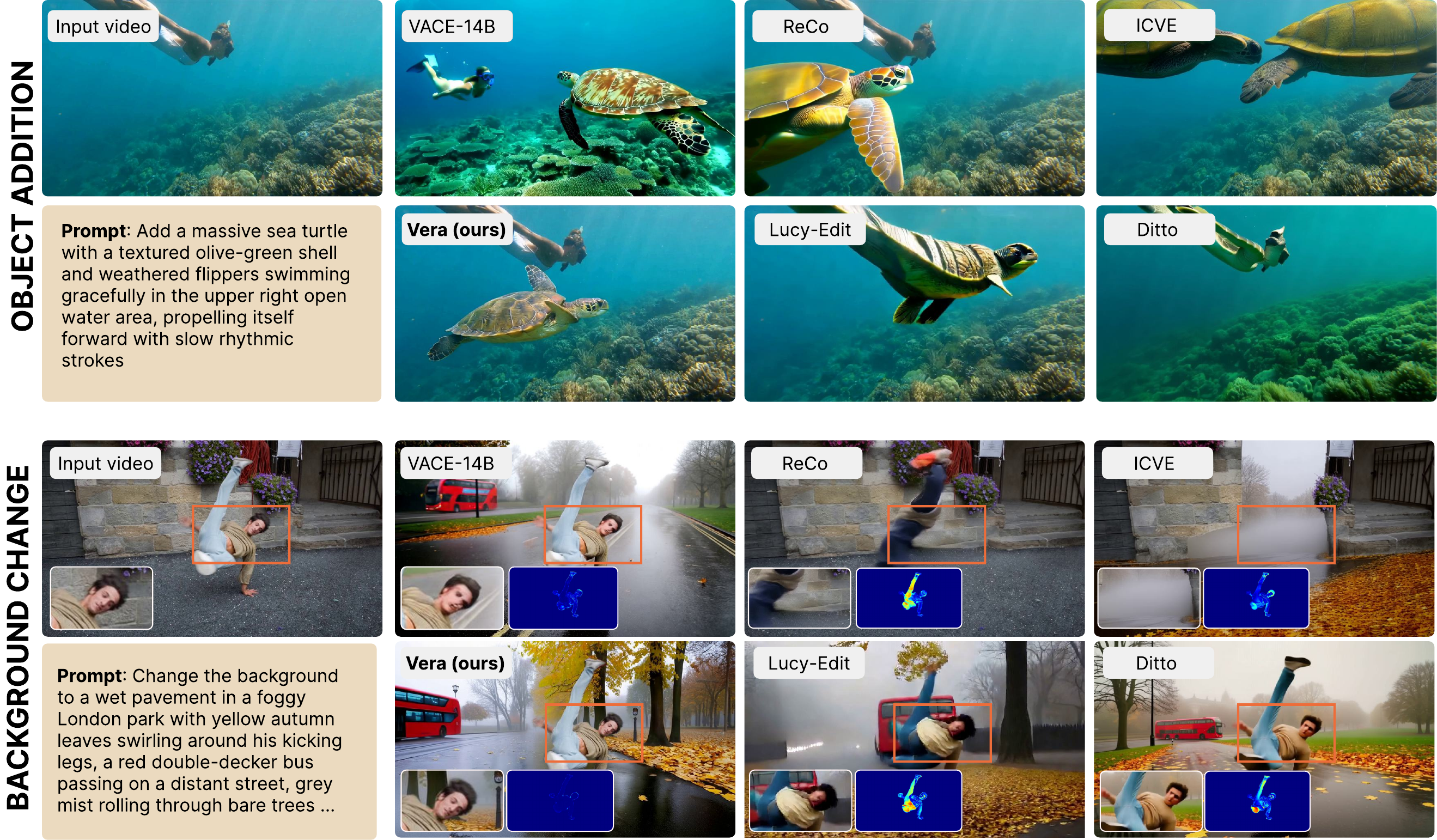}
  \caption{Qualitative comparison with existing video editing methods on background change (top) and object addition (bottom). For the background change example, each method includes sub-panels showing a zoom-in view (left) and a difference heat map over the preserved content region (right). Compared to end-to-end baselines that regenerate the entire video and introduce unintended changes to unedited regions, Vera preserves the original content best while maintaining high-quality edits.}
  \label{fig:visual-compare}
\end{figure}

\begin{figure}[ht]
  \centering
  \includegraphics[width=0.99\linewidth]{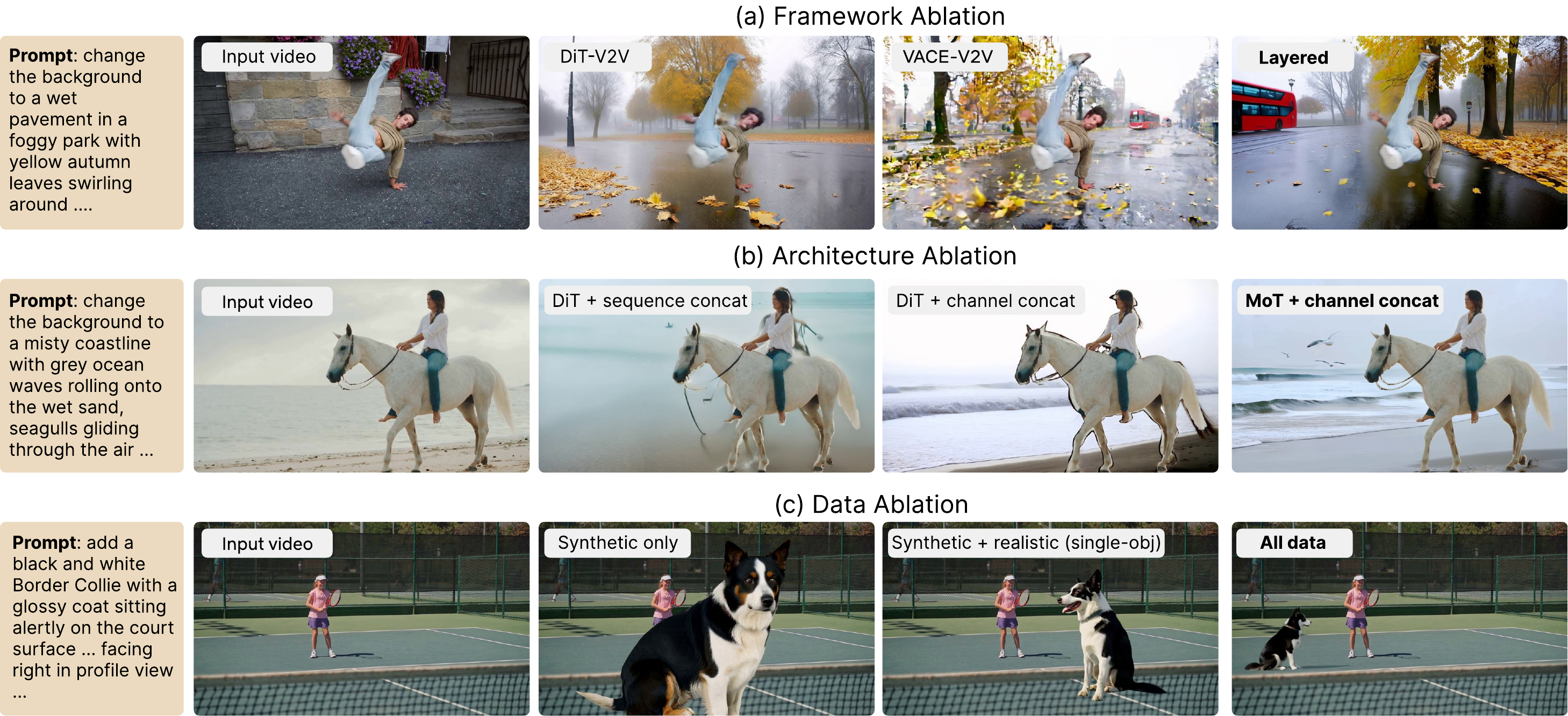}
  \caption{Qualitative examples from the ablation studies. Each row demonstrates the qualitative impact of a single design choice or training data variation while keeping all other variables fixed. \textbf{(a):} Layered editing paradigm (Vera) vs.\ standard video-to-video (V2V) architectures; zoom in to view the dancer's face. \textbf{(b):} Architecture choices within the layered framework, varying DiT design (Dense DiT vs.\ MoT) and input video conditioning. \textbf{(c):} Cumulative effect of each training data source. \textbf{Bold} column headers indicate the choices adopted in the final Vera model.}
  \label{fig:ablation-visual}
\end{figure}
 
We evaluate Vera on two representative video editing tasks -- background change and object addition -- to assess its content preservation, video quality, and instruction compliance.
Sec.~\ref{sec:setup} describes the experimental setup, including training details, evaluation protocol, and baselines. 
Sec.~\ref{sec:comparison} presents both qualitative and quantitative comparisons against existing methods, followed by a human preference study in Sec.~\ref{sec:user-study}.
Sec.~\ref{sec:ablation} ablates architectural design choices and the impact of different training data.

\subsection{Experimental Setup}\label{sec:setup}
\subsubsection{Training}
We train two model variants: Vera-1.3B and Vera-14B. Vera-1.3B initializes each of its three DiTs from Wan2.1-1.3B text-to-video model~\cite{wan2025wan}, yielding 3.9B parameters in total. 
Vera-14B initializes each DiT from Wan2.1-14B, yielding 42B parameters in total.
We train for 8{,}000 steps with a batch size of 16 at $832\times480$ resolution and 49 frames per clip. 
During training, we randomly drop the mask input so that both models learn to operate with and without mask video input.

\subsubsection{Evaluation}
We curate 72 test video--prompt pairs for object addition and 69 for background change, sourced from Pexels~\cite{pexels}, DAVIS~\cite{Pont-Tuset_arXiv_2017,Caelles_arXiv_2019}, and VACEBench~\cite{jiang2025vace}. The test videos span multiple difficulty levels, including slow to fast motion, various camera motions, single and multiple objects, and both simple and complex scenes.
For background change, we supply a coarse object mask video obtained from SAM~2~\cite{carion2025sam} to models that support mask input.
For object addition, no input mask is provided; models must determine the placement and extent of the inserted object themselves.
We evaluate along three complementary dimensions (Table~\ref{tab:eval-metrics}):
(1)~\textbf{Content preservation} measures whether regions outside the edit remain unaltered via pixel-level and perceptual similarity. For background change, similarity is computed over the preserved region annotated with alpha mattes; for object addition, it is computed on the full frame.
(2)~\textbf{Video quality} assesses temporal coherence and per-frame spatial quality.
(3)~\textbf{Instruction compliance} measures whether the edited video faithfully executes the editing prompt.
We adopt temporal flickering (TF), overall semantic consistency (OSC), and instruction satisfaction (IS) from VBench~\cite{huang2023vbench} and IVEBench~\cite{chen2026ivebench}, and report overall consistency (OC) as the average of VBench's subject consistency and background consistency.
We observe that classic video quality metrics from prior benchmarks often fail to distinguish between recent models. 
We therefore introduce two VLM-based video quality metrics---composition spatial quality (CS) and composition temporal quality (CT), which we find to be better aligned with human preference.
For all VLM-based metrics (CS, CT, and IS), we report the average score across three VLMs using identical system prompts: Gemini-3-Pro~\cite{google2026gemini3pro}, GPT-5.2~\cite{openai2026gpt52}, and Claude Sonnet-4.6~\cite{anthropic2026sonnet46}. Gemini-3-Pro receives native video input, while GPT-5.2 and Claude Sonnet-4.6 receive 32 uniformly sampled frames. 
System prompts for CS and CT are provided in the Appendix (Fig.~\ref{fig:composition-quality-prompt}).

\subsubsection{Baselines} We compare against seven recent open-source video editing models spanning four categories.
\textbf{General instruction-based models.} Ditto~\cite{bai2025scaling}, Lucy-edit~\cite{decart2025lucyedit}, and ICVE~\cite{liao2025context} treat video editing as direct video-to-video translation conditioned on a text instruction.
\textbf{Region-constrained editing.} ReCo~\cite{zhang2025reco} introduces regional constraints during training to improve consistency in non-edited regions.
\textbf{Mask-conditioned editing.} VACE~\cite{jiang2025vace} (1.3B and 14B) is an all-in-one video creation and editing framework that supports both video-to-video translation and mask-conditioned editing. VideoPainter~\cite{bian2025videopainter} is a mask-conditioned video inpainting model.
\textbf{Layered generation.} LayerFlow~\cite{ji2025layerflow} releases separate models for different tasks; we adopt its foreground-conditioned background generation model for background change only. 
Note that LayerFlow is hardcoded to generate 16 frames at $720\times 480$ resolution, so we report its performance on 16-frame sequences.

\subsection{Comparison with Existing Methods}\label{sec:comparison}

\subsubsection{Quantitative Results}
As reported in Table~\ref{tab:main-comparison}, Vera achieves substantially better content preservation than all baselines on both tasks. 
Vera-1.3B surpasses the strongest baseline by 3.5\,dB PSNR on background change and by 6.3\,dB on object addition,
while reducing structural error ($1{-}\text{SSIM}$) and perceptual distance (LPIPS) by more than half. 
Vera-14B extends these gains further, to 4.5\,dB and 7.1\,dB respectively. 
These improvements follow directly from the layered design, which preserves untouched content by construction, combined with the model's
accurate alpha matte prediction.
Traditional video quality metrics (OC, TF) are nearly saturated across all methods and provide little discrimination. 
The VLM-judged composition metrics reveal a more nuanced picture. 
On background change, Vera ranks second among all methods on composition quality. 
VACE achieves the highest CS, CT, and IS scores.
On object addition, Vera leads on both composition quality (CS, CT) and instruction compliance (IS).
Vera also achieves the highest IS on object addition, indicating that the layered design does not compromise edit faithfulness.

\subsubsection{User Study}\label{sec:user-study}
To complement the automated metrics, we conduct a human preference study using a standard two-alternative forced choice (2AFC) protocol. 
For each trial, annotators are shown the source video, the editing instruction, and two edited videos (Vera-1.3B vs.\ a randomly sampled baseline) displayed side by side with randomized left/right placement. 
Annotators answer three forced-choice questions, one per evaluation dimension: (1)~content preservation: which video better preserves the original content that should remain unchanged; 
(2)~video quality: which video has better overall visual quality and temporal consistency; 
and (3)~instruction compliance: which video more faithfully follows the editing instruction. 
We recruit 19 annotators, each assigned 32 pairwise trials. We exclude 2 annotators who returned fewer than 10 preference answers, and account for a small number of incomplete trials among the remaining annotators, yielding 513 valid trials in total (see Appendix~\ref{sec:appendix-user-study} for details).

As shown in Fig.~\ref{fig:user-study-winrates}, Vera-1.3B is preferred over all baselines on content preservation and instruction compliance.  
Video quality win rates are broadly consistent with the automated metrics, with the exception that Vera-1.3B is less preferred than VACE-14B and Ditto on background change video quality.
These results are achieved using 486K frames of layered training data.

\subsubsection{Qualitative Comparisons}
Qualitative comparisons shown in Fig.~\ref{fig:visual-compare} highlight two systematic failure modes in baseline methods that the quantitative metrics
capture only partially. 
In background change, V2V baselines introduce face distortion and body morphing in regions that should remain untouched.
In object addition, several baselines erroneously merge the inserted object with an existing foreground subject (\eg, fusing the sea turtle with the swimmer) or bleed its attributes (the olive-green color) into the surrounding background.
Vera avoids both failure modes: the alpha matte confines all generated content to the edit layer, leaving preserved regions identical to the input video. Beyond compositing, Vera's predicted alpha mattes are competitive with dedicated video matting methods on YouTubeMatte despite receiving no specialized matting loss or matting-only training (Appendix~\ref{sec:appendix-alpha-matte}).

\subsection{Ablation Study}\label{sec:ablation}
Our ablations isolate where Vera's content preservation and composition come from, holding the base model, training data, and budget fixed and varying one factor at a time: the layered paradigm improves preservation while retaining competitive composition (Sec.~\ref{sec:ablation-v2v}); the composite branch secures composition quality (Sec.~\ref{sec:ablation-composite}); the MoT architecture outperforms a dense DiT (Sec.~\ref{sec:ablation-arch}); and the alpha and composite branches benefit from faster learning rates (Sec.~\ref{sec:ablation-lr}).
Unless otherwise noted, all ablations use the Vera-1.3B model at step 6{,}000, with CS, CT, and IS averaged over three VLMs.

\begin{table}[t]
  \centering
  \caption{Layered editing substantially improves content preservation over standard video-to-video (V2V); full Vera remains competitive in composition and instruction-following (CS, CT, IS).
  All models are trained on the same data for the same training steps. 
  The model-name suffix denotes the nominal base-model size, while \#Params is the actual total. Vera is a mixture of three DiTs and Vera-no-comp drops the composite branch (2.6B). VLM-based metrics (CS, CT, IS) are averaged over three VLMs. Best results in \textbf{bold}, second-best \underline{underlined}.}
  \label{tab:ablation-arch-design}
  \small
  \resizebox{\linewidth}{!}{
  \begin{tabular}{lcccccccccc}
  \toprule
  \multirow{2}{*}{Design} & \multirow{2}{*}{\#Params}
    & \multicolumn{3}{c}{\shortstack{Content\\Preservation}}
    & \multicolumn{4}{c}{\shortstack{Video\\Quality}}
    & \multicolumn{2}{c}{\shortstack{Instruction\\Compliance}} \\
  \cmidrule(lr){3-5} \cmidrule(lr){6-9} \cmidrule(lr){10-11}
    & & PSNR $\uparrow$ & SSIM $\uparrow$ & LPIPS $\downarrow$
    & OC $\uparrow$ & TF $\uparrow$ & CS $\uparrow$ & CT $\uparrow$
    & OSC $\uparrow$ & IS $\uparrow$ \\
  \midrule
  \multicolumn{11}{l}{\textit{Object Addition}} \\
  \midrule
  V2V-DiT-1.3B & 1.4B & 21.1 & 0.861 & 0.210 & 0.98 & 0.98 & 3.29 & 3.26 & 0.243 & 3.90 \\
  V2V-VACE-1.3B & 2.2B & 20.4 & 0.861 & \underline{0.209} & 0.98 & 0.98 & \textbf{3.54} & \textbf{3.53} & \textbf{0.245} & \textbf{4.09} \\
  V2V-DiT-5B & 5.0B & 20.0 & 0.837 & 0.225 & 0.97 & 0.98 & 2.86 & 2.66 & \underline{0.244} & 3.66 \\
  V2V-VACE-5B & 7.6B & 21.3 & 0.856 & 0.219 & 0.97 & 0.98 & 2.98 & 2.71 & \underline{0.244} & 3.81 \\
  Vera-no-comp & 2.6B & \textbf{25.4} & \textbf{0.942} & \textbf{0.084} & 0.98 & 0.98 & 2.85 & 2.74 & \underline{0.244} & 3.54 \\
  Vera-1.3B (ours) & 3.9B & \underline{24.8} & \underline{0.941} & \textbf{0.084} & 0.98 & 0.98 & \underline{3.46} & \underline{3.48} & 0.243 & \underline{3.97} \\
  \midrule
  \multicolumn{11}{l}{\textit{Background Change}} \\
  \midrule
  V2V-DiT-1.3B & 1.4B & 31.0 & 0.979 & 0.026 & 0.96 & 0.97 & \underline{3.14} & \underline{3.09} & \textbf{0.243} & \textbf{4.35} \\
  V2V-VACE-1.3B & 2.2B & 31.0 & 0.980 & 0.025 & 0.96 & 0.97 & 3.07 & 3.05 & \underline{0.242} & \underline{4.34} \\
  V2V-DiT-5B & 5.0B & 29.7 & 0.972 & 0.032 & 0.96 & 0.98 & 2.88 & 2.54 & \textbf{0.243} & 4.21 \\
  V2V-VACE-5B & 7.6B & 30.3 & 0.976 & 0.029 & 0.96 & 0.97 & 2.21 & 2.04 & 0.240 & 3.75 \\
  Vera-no-comp & 2.6B & \textbf{35.0} & \textbf{0.992} & \textbf{0.010} & 0.96 & 0.97 & 2.85 & 2.68 & \underline{0.242} & 4.19 \\
  Vera-1.3B (ours) & 3.9B & \underline{33.8} & \underline{0.990} & \underline{0.012} & 0.96 & 0.97 & \textbf{3.24} & \textbf{3.24} & \underline{0.242} & 4.33 \\
  \bottomrule
  \end{tabular}
  }
  \end{table}

  \begin{table}[ht!]
  \centering
  \caption{Within-framework ablation on architecture choices for layered generation. We vary two axes: (1) DiT design -- a single DiT with all layer tokens concatenated in one sequence vs.\ an MoT architecture with separate DiTs per layer and joint self-attention; and (2) input video conditioning -- channel concatenation (with zero or copy initialization of the input video patch embedding) vs.\ sequence concatenation. All models evaluated at step 6{,}000. VLM-based metrics (CS, CT, IS) are averaged over three VLMs. Best results in \textbf{bold}, second-best \underline{underlined}.}
  \label{tab:ablation-layer-arch}
  \small
  \ra{0.95}
  \resizebox{\linewidth}{!}{
  \begin{tabular}{llccccccccc}
  \toprule
  \multirow{2}{*}{Architecture} & \multirow{2}{*}{Concat dim}
    & \multicolumn{3}{c}{\shortstack{Content\\Preservation}}
    & \multicolumn{4}{c}{\shortstack{Video\\Quality}}
    & \multicolumn{2}{c}{\shortstack{Instruction\\Compliance}} \\
  \cmidrule(lr){3-5} \cmidrule(lr){6-9} \cmidrule(lr){10-11}
    & & PSNR $\uparrow$ & SSIM $\uparrow$ & LPIPS $\downarrow$
    & OC $\uparrow$ & TF $\uparrow$ & CS $\uparrow$ & CT $\uparrow$
    & OSC $\uparrow$ & IS $\uparrow$ \\
  \midrule
  \multicolumn{11}{l}{\textit{Object Addition}} \\
  \midrule
  Dense DiT & Channel & 21.7 & 0.806 & 0.269 & 0.96 & 0.976 & 1.61 & 1.79 & 0.231 & 2.17 \\
  Dense DiT & Sequence  & 20.3 & 0.767 & 0.343 & 0.95 & \underline{0.978} & 1.32 & 1.46 & 0.223 & 1.75 \\
  MoT & Sequence & 23.5 & 0.920 & 0.103 & 0.98 & \textbf{0.979} & 3.40 & 3.47 & \underline{0.243} & 3.94 \\
  MoT & Channel-copy & \underline{23.9} & \underline{0.937} & \underline{0.090} & 0.98 & \textbf{0.979} & \textbf{3.49} & \textbf{3.50} & \textbf{0.244} & \textbf{4.01} \\
  MoT & Channel-zero & \textbf{24.8} & \textbf{0.941} & \textbf{0.084} & 0.98 & \underline{0.978} & \underline{3.46} & \underline{3.48} & \underline{0.243} & \underline{3.97} \\
  \midrule
  \multicolumn{11}{l}{\textit{Background Change}} \\
  \midrule
  Dense DiT & Channel & 31.6 & 0.987 & 0.015 & 0.96 & \textbf{0.977} & 2.49 & 2.57 & 0.240 & 3.91 \\
  Dense DiT & Sequence  & 32.4 & 0.989 & 0.014 & 0.95 & \underline{0.971} & 2.14 & 2.26 & 0.229 & 3.06 \\
  MoT & Sequence & \textbf{34.4} & \textbf{0.991} & \textbf{0.011} & 0.96 & 0.969 & \textbf{3.30} & \textbf{3.27} & \textbf{0.243} & \textbf{4.38} \\
  MoT & Channel-copy  & \underline{33.8} & \underline{0.990} & \underline{0.012} & 0.96 & 0.966 & 3.20 & 3.16 & \underline{0.242} & 4.32 \\
  MoT & Channel-zero & \underline{33.8} & \underline{0.990} & \underline{0.012} & 0.96 & \underline{0.971} & \underline{3.24} & \underline{3.24} & \underline{0.242} & \underline{4.33} \\
  \bottomrule
  \end{tabular}
  }
  \end{table}

\begin{table}[ht!]
  \centering
  \caption{Data ablation: cumulative effect of each training data source. Synthetic denotes VideoMatte240K~\cite{lin2021real} composites; +\,Single-obj adds realistic videos with a single prominent object; +\,Multi-obj further adds realistic videos with multiple objects. All models evaluated at step 6{,}000. VLM-based metrics (CS, CT, IS) are averaged over three VLMs. Best results in \textbf{bold}, second-best \underline{underlined}.}
  \label{tab:ablation-data}
  \small
  \ra{1.0}
  \resizebox{\linewidth}{!}{
  \begin{tabular}{lccccccccc}
  \toprule
  \multirow{2}{*}{Training data}
    & \multicolumn{3}{c}{\shortstack{Content\\Preservation}}
    & \multicolumn{4}{c}{\shortstack{Video\\Quality}}
    & \multicolumn{2}{c}{\shortstack{Instruction\\Compliance}} \\
  \cmidrule(lr){2-4} \cmidrule(lr){5-8} \cmidrule(lr){9-10}
    & PSNR $\uparrow$ & SSIM $\uparrow$ & LPIPS $\downarrow$
    & OC $\uparrow$ & TF $\uparrow$ & CS $\uparrow$ & CT $\uparrow$
    & OSC $\uparrow$ & IS $\uparrow$ \\
  \midrule
  \multicolumn{10}{l}{\textit{Object Addition}} \\
  \midrule
  Synthetic only & 18.4 & \underline{0.823} & 0.199 & 0.97 & 0.976 & 2.57 & 2.81 & 0.240 & 3.33 \\
  +\,Realistic (single-obj) & \underline{19.1} & 0.817 & \underline{0.198} & \textbf{0.98} & \textbf{0.978} & \underline{3.17} & \underline{3.27} & \underline{0.241} & \underline{3.79} \\
  +\,Realistic (multi-obj) & \textbf{24.8} & \textbf{0.941} & \textbf{0.084} & \textbf{0.98} & \textbf{0.978} & \textbf{3.46} & \textbf{3.48} & \textbf{0.243} & \textbf{3.97} \\
  \midrule
  \multicolumn{10}{l}{\textit{Background Change}} \\
  \midrule
  Synthetic only & \underline{33.8} & \underline{0.990} & \underline{0.012} & 0.96 & \textbf{0.977} & 3.21 & \underline{3.28} & 0.241 & 4.28 \\
  +\,Realistic (single-obj) & \textbf{35.6} & \textbf{0.993} & \textbf{0.010} & 0.96 & \underline{0.975} & \textbf{3.26} & \textbf{3.32} & \textbf{0.242} & \textbf{4.39} \\
  +\,Realistic (multi-obj) & \underline{33.8} & \underline{0.990} & \underline{0.012} & 0.96 & 0.971 & \underline{3.24} & 3.24 & \textbf{0.242} & \underline{4.33} \\
  \bottomrule
  \end{tabular}
  }
  \end{table}
\subsubsection{Video-to-video vs.\ layered editing}
\label{sec:ablation-v2v}
Layered editing substantially improves content preservation while retaining competitive composition and edit quality.
With the training data and gradient steps held fixed (architectural differences in Fig.~\ref{fig:architecture}), Vera and Vera-no-comp outperform the 1.3B V2V baselines on preservation by 2.8--5.0\,dB PSNR (Table~\ref{tab:ablation-arch-design}). 
Full Vera also remains competitive with V2V on composition (CS, CT) and instruction-following.
This gap follows from the output representation: V2V models regenerate the full video, allowing changes to propagate beyond the intended edit, 
whereas Vera explicitly isolates the generative edit in the RGBA representation and directly retains source pixels outside the predicted alpha support.
Accordingly, the qualitative results show distortions in regions regenerated by V2V that remain unchanged in Vera's output (Fig.~\ref{fig:ablation-visual}, top row).
The Wan2.2-5B results provide an additional comparison with larger V2V models that also fail to close the gap, but do not constitute a controlled scaling study because they use a different base model.

\subsubsection{Effect of the composite layer}
\label{sec:ablation-composite}
The composite branch is what keeps Vera's composition competitive with end-to-end V2V.
Removing it (Vera-no-comp) improves raw preservation slightly, but drops composition and instruction compliance below even the 1.3B V2V baselines: on object addition, CS falls from 3.46 to 2.85 and IS from 3.97 to 3.54 (Table~\ref{tab:ablation-arch-design}).
Trained on the natural-video distribution, the composite layer regularizes the edit and alpha branches and improves cross-layer harmonization. The three-output formulation also provides the variables needed in principle to recover large semi-transparent preserved regions that a two-layer model cannot resolve, but our experiments use the $V_{\mathrm{preserved}}\approx V_{\mathrm{src}}$ approximation and do not evaluate this case (Remark~\ref{remark:semi-transparent}).

\subsubsection{Architecture ablation}
\label{sec:ablation-arch}
Table~\ref{tab:ablation-layer-arch} ablates two orthogonal design axes within the layered framework. The first axis is the DiT design: a single DiT that concatenates all layer tokens (edit, alpha, composite) into one sequence versus an MoT architecture with separate DiTs per layer and joint self-attention. The second axis is input video conditioning: channel concatenation versus sequence concatenation.
For channel concatenation, we use separate patch embedding layers for the composite layer video and the input video. The composite layer's embedding is initialized from pretrained weights; the input video embedding is either zero-initialized (channel-zero) or initialized by copying the pretrained weights (channel-copy).
The MoT design is clearly superior to a single dense DiT across all three dimensions. The second row of Fig.~\ref{fig:ablation-visual} further reveals that MoT greatly improves alignment between the edit layer and the alpha matte, leading to better compositing. We also find that sequence concatenation yields better results on background change while channel concatenation is better for object addition. Since sequence concatenation incurs substantially higher FLOPs, we adopt channel concatenation for our main experiments.

\subsubsection{Per-layer learning rate}
\label{sec:ablation-lr}
Table~\ref{tab:ablation-lr} investigates the effect of assigning different learning rates to the three DiT branches using Vera-1.3B. Since the edit layer, alpha matte, and composite video have substantially different distributions, each branch may benefit from a tailored learning rate. We fix a base learning rate of $10^{-5}$ and express each branch's rate as a multiple of this base. Starting from a uniform rate (1$\times$ for all branches), increasing the alpha branch to 10$\times$ (Config~A) notably improves content preservation on background change (+2.5\,dB PSNR) and instruction compliance on object addition (IS: 3.00$\to$4.00). Further increasing the composite branch to 10$\times$ (Config~B) yields substantial gains on object addition content preservation (+3.2\,dB PSNR over Config~A). Finally, reducing the edit branch to 0.1$\times$ (Vera's default) yields comparable overall performance to Config~B, with no clear winner between the two. We adopt the 0.1$\times$ edit learning rate and fix it throughout all other experiments. Overall, the results indicate that the alpha and composite branches benefit from faster adaptation relative to the edit branch.

\begin{table}[ht!]
\centering
\caption{Per-layer learning rate ablation. Each DiT branch is assigned a learning rate as a multiple of the base rate $10^{-5}$. Config-A increases alpha branch base learning rate by 10$\times$. Config-B further increases composite branch learning rate by 10$\times$. The highlighted row is Vera's default configuration. VLM-based metrics (CS, CT, IS) are averaged over three VLMs.}
\label{tab:ablation-lr}
\small
\resizebox{\linewidth}{!}{
\begin{tabular}{lcccccccccccc}
\toprule
\multirow{2}{*}{Config}
  & \multicolumn{3}{c}{Learning rate}
  & \multicolumn{3}{c}{\shortstack{Content\\Preservation}}
  & \multicolumn{4}{c}{\shortstack{Video\\Quality}}
  & \multicolumn{2}{c}{\shortstack{Instruction\\Compliance}} \\
\cmidrule(lr){2-4} \cmidrule(lr){5-7} \cmidrule(lr){8-11} \cmidrule(lr){12-13}
  & Edit & Alpha & Comp.
  & PSNR $\uparrow$ & SSIM $\uparrow$ & LPIPS $\downarrow$
  & OC $\uparrow$ & TF $\uparrow$ & CS $\uparrow$ & CT $\uparrow$
  & OSC $\uparrow$ & IS $\uparrow$ \\
\midrule
\multicolumn{13}{l}{\textit{Object Addition}} \\
\midrule
Uniform           & 1x   & 1x  & 1x  & 23.1 & 0.871 & 0.160 & 0.98 & 0.98 & 2.30 & 2.45 & 0.237 & 3.00 \\
Config-A          & 1x   & 10x & 1x  & 22.8 & 0.929 & 0.102 & 0.98 & 0.98 & 3.29 & 3.41 & 0.243 & 4.00 \\
Config-B          & 1x   & 10x & 10x & 26.0 & 0.956 & 0.072 & 0.98 & 0.98 & 3.36 & 3.48 & 0.243 & 4.00 \\
\rowcolor{gray!20}
Vera              & 0.1x & 10x & 10x & 24.8 & 0.941 & 0.084 & 0.98 & 0.98 & 3.46 & 3.48 & 0.243 & 3.97 \\
\midrule
\multicolumn{13}{l}{\textit{Background Change}} \\
\midrule
Uniform           & 1x   & 1x  & 1x  & 32.2 & 0.989 & 0.014 & 0.96 & 0.97 & 2.98 & 3.06 & 0.242 & 4.31 \\
Config-A          & 1x   & 10x & 1x  & 34.7 & 0.991 & 0.011 & 0.96 & 0.97 & 3.08 & 3.13 & 0.241 & 4.34 \\
Config-B          & 1x   & 10x & 10x & 33.6 & 0.990 & 0.012 & 0.96 & 0.97 & 3.29 & 3.29 & 0.243 & 4.37 \\
\rowcolor{gray!20}
Vera              & 0.1x & 10x & 10x & 33.8 & 0.990 & 0.012 & 0.96 & 0.97 & 3.24 & 3.24 & 0.242 & 4.33 \\
\bottomrule
\end{tabular}
}
\end{table}

\subsubsection{Data ablation}
We present quantitative and qualitative study on the contribution of training data source in Table~\ref{tab:ablation-data} and the third row of Fig.~\ref{fig:ablation-visual}, respectively. 
The synthetic layered subset provides accurate alpha supervision, including fine structures such as hair, but is limited to centered human subjects with static cameras. 
However, as this subset lacks diversity in object scale, camera motion, and scene layout, the model receives no supervision for matching these properties and struggles to generalize to in-the-wild videos.
Adding realistic single-object data introduces diverse, dynamic scenes that substantially boost background change preservation (PSNR: 33.8$\to$35.6\,dB) and improve composition quality (CS, CT) and instruction compliance (IS) on both tasks. The third dataset targets object addition with complex, multi-object scenes and interactive effects; it yields a large jump in object addition preservation (PSNR: 19.1$\to$24.8\,dB) with some regression on background change, suggesting that balancing the data mixture across tasks remains an important direction. 
The dog example in Figure 8 reflects these improvements, from having no interactions and occupying at the very front of the entire video scene with synthetic-only data, to slow gaining realistic effects such as shadows and finally becoming the right size when all data are utilized.

\section{Conclusion and Limitations}

We investigated how to introduce editable layer structure into diffusion models for video editing, where generated edit layers must support coherent compositing with the source video. 
Vera provides a concrete formulation: it jointly produces an edit layer, an alpha matte, and a composite video, separating what to generate from what to preserve. 
The resulting editable layers can support iterative refinement in downstream editing workflows.
Through controlled experiments, we identified three key ingredients that enable layer separation while retaining competitive composition quality: an MoT architecture with cross-layer interaction through joint self-attention, composite-branch supervision, and curated layered data with accurately aligned edit layers and alpha mattes.

Three limitations remain in this work. 
First, jointly generating three layers increases inference cost: Vera-1.3B is roughly $3\times$ slower than VACE (Appendix Table~\ref{tab:inference-cost}). 
Second, our evaluation is limited to object addition and background replacement. Extending the approach to relighting, complex visual effects, and broader editing operations will require layered supervision that captures the corresponding interactions. 
Third, our inference procedure approximates $V_{\mathrm{preserved}}$ with $V_{\mathrm{src}}$ and therefore assumes that preserved content contains only small semi-transparent regions (Remark~\ref{remark:semi-transparent}). Direct recovery in cases such as glass or water requires suitable layered training data and explicit evaluation. Addressing these boundaries would extend layered generation toward a broader set of production editing operations.

\bibliographystyle{arxiv/preprint}
\bibliography{main}

\appendix
\newpage
\section{Supplementary Material}

\subsection{Training data}

\subsubsection{Data pipeline overview}
\label{sec:appendix-data-pipeline}
As described in the main paper, our layered training data is constructed from two types of video sources: the video matting dataset~\cite{lin2021real}, and real-world videos collected from Pexels~\cite{pexels} and Mixkit~\cite{mixkit}. 
Both pipelines are multi-stage processes that combine automated tools (\eg, segmentation, matting, inpainting, and VLM-based captioning with Gemini-3-Pro) with human annotators, who identify suitable objects, provide point prompts to SAM2 for segmentation, and filter out low-quality samples at each stage. 

Fig.~\ref{fig:data-pipelines} illustrates the complete data construction pipelines for (a) object addition and (b) background change, showing the full processing chain starting from real internet videos. 
For the VideoMatte240K dataset, since it already provides high-quality alpha mattes with precise boundaries, the first two stages---video collection/filtering and matting---can be skipped, and the pipeline proceeds directly from the object removal stage onward. 
We describe each pipeline below. Table~\ref{tab:licenses} lists all data sources, models, and tools used in the data pipelines along with their licenses.

\paragraph{Object addition pipeline.}
The object addition pipeline (Fig.~\ref{fig:data-pipelines}a) consists of six stages:
\textbf{(1) Video collection and filtering.} We collect raw videos from Pexels~\cite{pexels} and Mixkit~\cite{mixkit}, then filter and preprocess them to obtain a set of high-quality videos with sufficient resolution and visual diversity, covering various scenes and objects.
\textbf{(2) Matting.} Human annotators identify suitable object(s) in each video and provide point prompts to SAM2~\cite{ravisam}. The SAM2 segmentation masks are then refined with VideoMaMa~\cite{lim2026videomama} to obtain high-quality alpha mattes with accurate boundaries.
\textbf{(3) Object removal.} Using the alpha mattes, we remove the selected object(s) from the video with a video object removal model (Casper-1.3B~\cite{lee2025generative}) to obtain clean background videos. Since the removal model can occasionally produce artifacts, this stage includes a filtering step to discard failed samples.
\textbf{(4) Omnimatte optimization.} For objects with associated visual effects (\eg, shadows and reflections), we apply omnimatte optimization~\cite{lee2025generative} to extract alpha mattes that capture not only the object itself but also its effects. This step is skipped for objects without associated effects.
\textbf{(5) Effect refinement.} The omnimatte optimization results are often noisy. Human annotators identify regions containing actual effects (as opposed to artifacts) and use SAM2 to segment these regions. The resulting mask is used to refine the alpha by zeroing out regions beyond the mask, producing clean edit layers and alpha mattes.
\textbf{(6) Captioning.} Finally, a VLM generates edit instructions describing the object addition (see Fig.~\ref{fig:obj-add-prompt} for the system prompt), producing the final (input video, edit instruction, edit layer, alpha matte) training tuple.

\paragraph{Background change pipeline.}
The background change pipeline (Fig.~\ref{fig:data-pipelines}b) shares the first two stages (video collection and matting) with the object addition pipeline---in practice, we often reuse the alpha matte results directly. It then diverges as follows:
\textbf{(3) Object removal.} As in the object addition pipeline, we remove the identified object(s) from the video using Casper-1.3B~\cite{lee2025generative}. However, unlike object addition, here we want the object's visual effects (\eg, shadows, reflections) to remain in the background, since the background with effects will serve as our edit layer.
\textbf{(4) Effects transfer.} We transfer pixels from the original video that lie beyond the alpha matte region back into the object-removed video, so that only the object itself is removed while its associated effects are preserved in the background. The resulting video---the original background with effects intact but the object removed---becomes the edit layer.
\textbf{(5) Background generation and recomposition.} We use a VLM to analyze the matted object(s) and generate scene prompts describing backgrounds that would naturally fit with them. These prompts are sent to VACE inpainting~\cite{jiang2025vace} to generate a synthetic background, which is then composited with the object using the alpha matte to produce the input video.
\textbf{(6) Captioning.} We first use a VLM to generate detailed captions of both the objects and the scene for the input video and the target composite using the system prompt in Fig.~\ref{fig:video-caption-prompt}. These captions are then provided to a separate VLM call (Fig.~\ref{fig:bg-change-prompt}) to infer the background change editing instruction.

\subsection{Model details}

Table~\ref{tab:model-comparison} summarizes the architecture configurations. Vera extends the base Wan2.1 T2V DiT into a MoT with three separate DiTs—one each for the edit layer, alpha matte, and composite video—that interact through joint self-attention. The number of transformer layers is inherited from the base model (30 for the 1.3B variant and 40 for the 14B variant). Because each DiT maintains its own set of parameters, the total parameter count is roughly three times that of the base T2V model. However, per-step FLOPs increase by more than 3$\times$ because joint self-attention operates over the combined token sequence from all three DiTs, and attention cost grows quadratically with sequence length.

\begin{table}[h]
  \centering
  \caption{Architecture comparison between the base Wan2.1 T2V models and the corresponding Vera variants. TFLOPs is measured per denoising step.}
  \label{tab:model-comparison}
  \small
  \ra{1.1}
  \begin{tabular}{lcccc}
  \toprule
  Model & \# Parameters & TFLOPs & Layers & Output Layers \\
  \midrule
  Wan2.1-T2V-1.3B & 1.4B & 129 & 30 & 1 \\
  Vera-1.3B       & 3.9B & 840 & 30 & 3 \\
  Wan2.1-T2V-14B  & 14B  & 832 & 40 & 1 \\
  Vera-14B        & 38.2B & 4,514 & 40 & 3 \\
  \bottomrule
  \end{tabular}
\end{table}

Vera-1.3B generates a video in roughly $8.3$\,min on a single A100 with $21.8$\,GB peak VRAM, about $3\times$ slower than VACE~\cite{jiang2025vace} (Table~\ref{tab:inference-cost}).
This overhead---from the joint self-attention over all three layers, which raises per-step FLOPs---remains practical and can be reduced with standard techniques such as kernel fusion and sequence parallelism.

\begin{table}[h]
  \centering
  \caption{Vera's layered generation is ${\sim}3\times$ slower than VACE~\cite{jiang2025vace} but remains practical. VRAM is peak reserved memory; FLOPs are per denoising step; Time is the total generation time per video.}
  \label{tab:inference-cost}
  \small
  \ra{1.1}
  \begin{tabular}{lcccc}
  \toprule
  Model & \#A100 & VRAM (GB) & FLOPs (TF) & Time (min) \\
  \midrule
  VACE-1.3B & 1 & 25.3 & 194 & 2.4 \\
  Vera-1.3B & 1 & 21.8 & 840 & 8.3 \\
  \midrule
  VACE-14B & 4 & 37.2 & 1{,}007 & 5.6 \\
  Vera-14B & 4 & 42.4 & 4{,}514 & 14.9 \\
  \bottomrule
  \end{tabular}
\end{table}

\paragraph{Training details}
We train Vera with Adam, using individual base learning rates for each DiT as determined by the per-layer learning-rate ablation in Table~\ref{tab:ablation-lr}. 
We apply a linear warmup of 200 steps. To encourage the model to generate both with and without the mask input condition, we randomly drop the mask input during training. 
We use a mask dropout rate of 0.3 for background change and 0.9 for object addition. The losses for the three output layers are equally weighted.
Each of the five input streams---edit layer latent, alpha layer latent, composite layer latent, mask video latent, and input video latent---has its own dedicated patch embedding layer. 
The patch embeddings for the mask video and input video are initialized from zero, while the remaining patch embedding layers are initialized from the pre-trained T2V model.

\subsection{Alpha matte quality}
\label{sec:appendix-alpha-matte}
Despite receiving no specialized matting loss or matting-only training, Vera's predicted alpha mattes are competitive with dedicated video matting methods on YouTubeMatte.
Table~\ref{tab:vera-youtubematte} evaluates the predicted mattes on the YouTubeMatte benchmark (resized to $832 \times 480$) against the matting specialist MatAnyone~\cite{yang2025matanyone}, the raw SAM2 mask~\cite{ravisam}, and VideoMaMa~\cite{lim2026videomama}.
Vera-14B is on par with MatAnyone and well above the SAM2 mask, trailing only VideoMaMa---which is expected, as VideoMaMa is itself used to produce the alpha supervision in our data pipeline (Sec.~\ref{sec:appendix-data-pipeline}) and thus acts as an effective upper bound.
This indicates that explicit alpha supervision within the layered editing objective can produce accurate mattes without specialized matting losses or matting-only training.

\begin{table}[h]
  \centering
  \caption{Without specialized matting losses or matting-only training, Vera-14B is competitive with dedicated matting methods on YouTubeMatte. Lower is better for all metrics; best in \textbf{bold}.}
  \label{tab:vera-youtubematte}
  \small
  \ra{1.1}
  \begin{tabular}{lccccc}
  \toprule
  Method & MSE$\downarrow$ & MAD$\downarrow$ & Grad$\downarrow$ & dtSSD$\downarrow$ & Conn$\downarrow$ \\
  \midrule
  VideoMaMa~\cite{lim2026videomama} & \textbf{0.364} & \textbf{1.595} & \textbf{0.984} & \textbf{1.201} & \textbf{0.551} \\
  MatAnyone~\cite{yang2025matanyone} & 0.856 & 2.464 & 2.303 & 1.705 & 0.970 \\
  SAM2 mask~\cite{ravisam} & 2.397 & 3.766 & 5.965 & 3.743 & 1.418 \\
  \midrule
  Vera-1.3B & 1.035 & 3.225 & 2.890 & 2.334 & 0.969 \\
  Vera-14B & 0.824 & 2.544 & 2.311 & 1.982 & 0.845 \\
  \bottomrule
  \end{tabular}
\end{table}

\subsection{User study}
\label{sec:appendix-user-study}

\begin{figure}[t]
  \centering
  \includegraphics[width=0.99\linewidth]{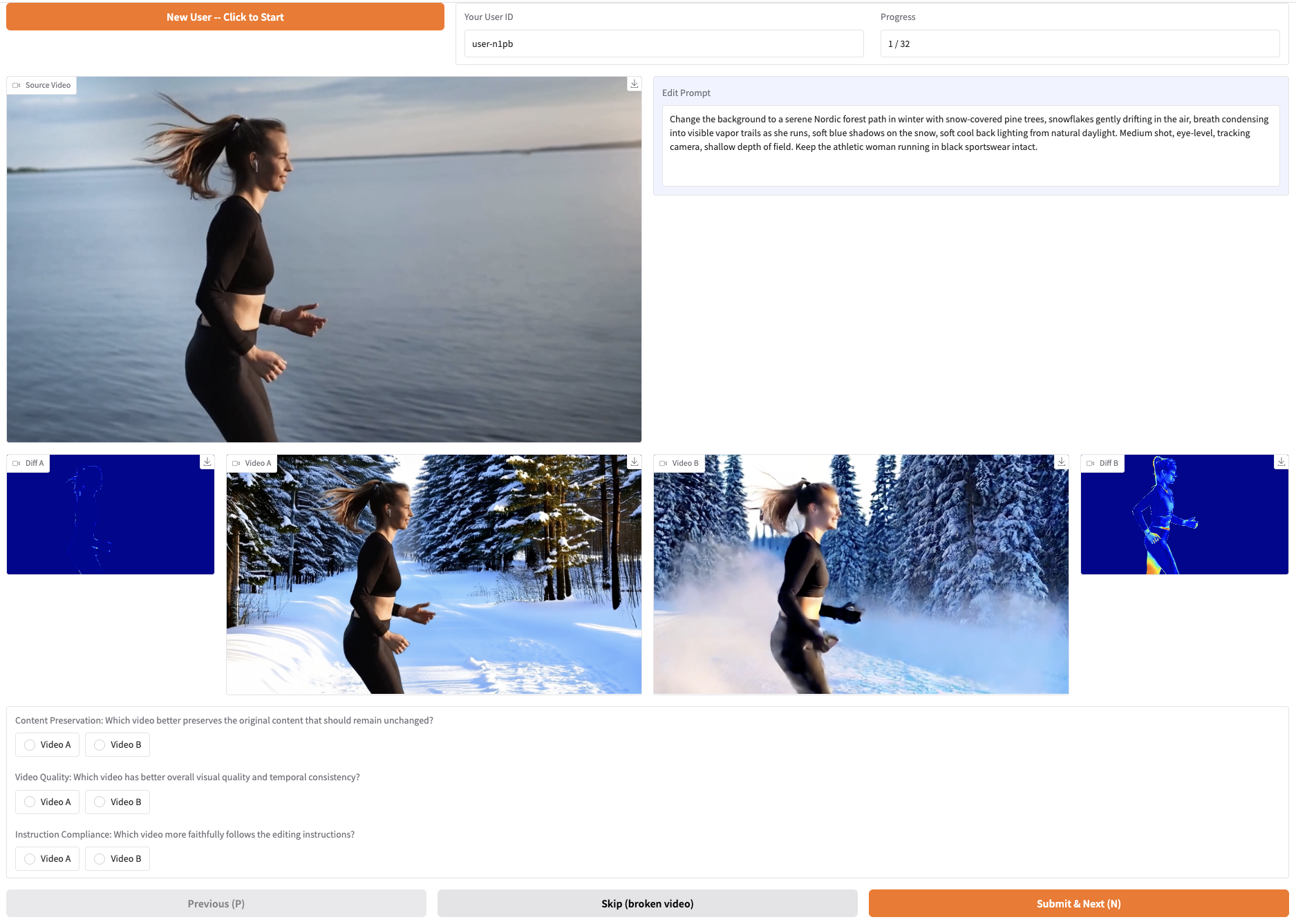}
  \caption{Screenshot of the annotation interface used in our human preference study. Each user is assigned an anonymous id. For each trial, annotators view the source video and the editing instruction at the top, followed by two anonymized edited videos (Video~A and~B) with their corresponding difference heatmaps (Diff~A and~B). Annotators select their preference for each of the three evaluation dimensions via forced-choice buttons.}
  \label{fig:user-study-ui}
\end{figure}

Fig.~\ref{fig:user-study-ui} shows the annotation interface used for our human preference study. 
As described in the main paper, we adopt a two-alternative forced choice (2AFC) protocol. 
Each trial presents the annotator with the source video (top left), the editing instruction (top right), and two anonymized edited videos (Video~A and Video~B) displayed side by side in the bottom row. 
The assignment of Vera and the baseline to the left or right position is randomized per trial to avoid positional bias. 
To aid content preservation judgments, difference heatmaps (Diff~A and Diff~B) are displayed alongside each video, highlighting pixel-level deviations from the source. 
Annotators then answer three forced-choice questions covering content preservation, video quality, and instruction compliance.
We recruited 19 annotators, each assigned 32 pairwise trials, for a maximum of $19 \times 32 = 608$ trials. 
We exclude the responses of annotators who returned fewer than 10 preference answers, which filters out 2 annotators and leaves 17 effective annotators ($17 \times 32 = 544$ trials).
Among the remaining annotators, a small number of trials were not completed due to technical interruptions (\eg, network connectivity issues) or annotator abstentions, yielding 513 valid trials in total.

\begin{table}[t]
    \centering
    \caption{Evaluation metrics grouped by dimension. Arrows indicate whether higher ($\uparrow$) or lower ($\downarrow$) is better. The system prompt for the VLM-judged CS and CT metrics is shown in Fig.~\ref{fig:composition-quality-prompt}.}
    \label{tab:eval-metrics}
    \small
    \begin{tabularx}{0.9\textwidth}{@{}llX@{}}
    \toprule
    Dimension & Metric & Description \\
    \midrule
    \multirow{3}{*}{\shortstack[l]{Content\\Preservation}}
      & PSNR $\uparrow$ & Peak signal-to-noise ratio \\
      & SSIM $\uparrow$ & Structural similarity \\
      & LPIPS $\downarrow$ & Learned perceptual similarity \\
    \midrule
    \multirow{4}{*}{\shortstack[l]{Video\\Quality}}
      & OC $\uparrow$ & Overall consistency: average of subject consistency (DINO~\cite{huang2023vbench}) and background consistency (CLIP~\cite{huang2023vbench}) across frames \\
      & TF $\uparrow$ & Temporal flickering: mean frame difference~\cite{huang2023vbench} \\
      & CS $\uparrow$ & Composition spatial quality: VLM-judged per-frame spatial blending, averaged over 3 VLMs (ours) \\
      & CT $\uparrow$ & Composition temporal quality: VLM-judged temporal coherence of the edit, averaged over 3 VLMs (ours) \\
    \midrule
    \multirow{2}{*}{\shortstack[l]{Instruction\\Compliance}}
      & OSC $\uparrow$ & Overall semantic consistency between edited video and target prompt via VideoCLIP-XL2~\cite{chen2026ivebench} \\
      & IS $\uparrow$ & Instruction satisfaction: VLM-judged accuracy on a 5-point scale, averaged over Gemini-3-Pro, GPT-5.2, and Claude Sonnet-4.6~\cite{chen2026ivebench} \\
    \bottomrule
    \end{tabularx}
\end{table}

\begin{table}[t]
\centering
\caption{Licenses for datasets, annotation tools, and foundation models used in this work.}
\label{tab:licenses}
\small
\begin{tabularx}{0.9\textwidth}{@{}p{0.5\textwidth}X@{}}
\toprule
Data / Model / Tool & License \\
\midrule
Pexels videos~\cite{pexels} & \href{https://www.pexels.com/license/}{Pexels Free License} \\
Mixkit videos~\cite{mixkit} & \href{https://mixkit.co/license/\#videoFree}{Mixkit Free License} \\
VideoMatte240K~\cite{lin2021real} & MIT License \\
DAVIS~\cite{Pont-Tuset_arXiv_2017,Caelles_arXiv_2019} & BSD 3-Clause License \\
VACEBench~\cite{jiang2025vace} & Apache-2.0 License \\
SAM2~\cite{ravisam} & Apache-2.0 License \\
VideoMama~\cite{lim2026videomama} & CC BY-NC 4.0 \\
VACE~\cite{jiang2025vace} & Apache-2.0 License \\
Minimax-Remover~\cite{zi2025minimax} & CC BY-NC 4.0 \\
Generative Omnimatte~\cite{lee2025generative} & Apache-2.0 License \\
Wan2.1-14B~\cite{wan2025wan} & Apache-2.0 License \\
\bottomrule
\end{tabularx}
\end{table}

\begin{figure}[t]
  \centering
  \begin{tcolorbox}[promptbox]
\begin{lstlisting}[style=prompttext]
You are an expert video editing instructor. You will receive two videos: a "before" video (clean background without the object) and an "after" video (with the object composited in). Your task is to generate a detailed editing instruction describing the added object.

Your Task
Compare the two videos carefully and write a thorough instruction describing everything about the added subject - its appearance, motion, and placement in the scene. Provide as much visual detail as the videos justify; do not artificially shorten or summarize your description.

Prompt Formula:
Add [detailed subject description] [action/motion with specifics] [precise location/placement in the scene]. [Camera and lighting constraints].

What to Describe

Subject Appearance
- Physical attributes: size, shape, build, posture
- Colors and textures: clothing materials, fur, skin tone, surface finish
- Distinctive features: accessories, patterns, markings, facial expression
- If the subject is a person: clothing details (garments, colors, fit), hair, visible accessories
- If the subject is an animal: breed/species cues, coat color and texture, size relative to scene
- If the subject is an object: material, color, condition, style

Motion and Action
- Primary action: what the subject is doing (e.g., walking, sitting, spinning, hovering)
- Motion direction and speed: left to right, toward camera, slowly, briskly
- Body dynamics: limb positions, posture changes, secondary motion (hair swaying, clothes fluttering)
- Interaction: if the subject interacts with anything already in the scene, describe the interaction

Spatial Placement
- Position in frame: left/center/right, foreground/midground/background
- Relative scale: how large the subject appears compared to the scene
- Facing direction - include for any subject that has a distinguishable front (people, animals, vehicles, aircraft, etc.). Omit only for symmetric objects with no clear front (e.g., a ball, a box, a cup):
  facing the camera: Front visible (face, headlights, nose cone, etc.)
  three-quarter view: Angled between front and side
  profile view (left/right): Pure side view
  facing away: Rear toward the camera (back of head, tail, tailgate, etc.)
  looking over the shoulder: Body facing away but head turned back (people/animals only)

- Depth placement: how far from the camera the subject appears

Lighting and Shadow Interaction
- If visible: describe any shadows cast by the added subject
- Describe how the lighting falls on the subject (lit from which side, rim-lit, silhouetted, etc.)

Camera Constraints
- Shot size as it relates to the subject: how much of the subject is visible (full body, waist up, close-up, etc.)
- Camera angle relative to the subject: eye-level, low angle, high angle, etc.
- Background focus around the subject: blurred background, sharp background, soft bokeh, etc.

DO:
- Start with "Add"
- Describe only what is clearly visible - be specific and concrete
- Include all visually relevant details about the added subject
- Always include facing direction for people and animals (e.g., facing the camera, facing away, profile view, three-quarter view). This is mandatory - do not omit it
- Include camera constraints: shot size, camera angle, and background focus
- Keep the instruction under 80 words - prioritize the most visually important details
- Describe motion over time if it changes across the video

DON'T:
- Do NOT describe the background or environment - only describe the added object and its interaction with the scene
- Do NOT reference what was in the "before" video or what the object replaces - describe the added subject as if it simply appears in the scene
- Do NOT hallucinate details that are not clearly visible in the videos
- Do NOT use vague descriptors like "nice", "interesting"

Output Format
Return a JSON object with a single "instruction" field containing the editing instruction.

Examples
Output: {"instruction": "Add a yellow taxi cab parked diagonally along the left side of the street, front bumper angled toward the camera, glossy paint reflecting the overcast sky, license plate partially visible, faint shadow cast underneath onto the wet asphalt. Wide shot, eye-level, sharp background."}

Output: {"instruction": "Add a young child in a royal-blue puffer jacket and dark jeans sitting cross-legged on the right end of the wooden bench, facing the camera, leaning slightly forward with hands on their knees, quilted texture catching overhead light with small highlights along the seams. Medium shot, slightly high angle, slightly blurred background."}

Output: {"instruction": "Add a white domestic shorthair cat walking steadily from right to left along the windowsill in the foreground, facing left in profile, tail held upright curving at the tip, each paw placed precisely on the narrow ledge, soft side-lighting from the window on the left side of its body. Medium close-up, eye-level, dreamy out-of-focus background."}
\end{lstlisting}
  \end{tcolorbox}
  \caption{The system prompt used for generating edit instructions for the object addition data.}
  \label{fig:obj-add-prompt}
\end{figure}

\begin{figure}[t]
  \centering
  \begin{tcolorbox}[promptbox]
\begin{lstlisting}[style=prompttext]
You are an expert cinematographer and video description AI. Your task is to write a concise, accurate caption for a given video, suitable for use as a prompt for AI video diffusion models.

Input: You will receive a video. Watch the video and write a high-quality prompt to describe the video.

Output Format: Write a high-quality prompt following this formula:

[Subject(s) description] + [action/motion] + [environment/scene] + [Constraints: lighting/color description + shot/camera description]

For subject(s), action, and environment: compose natural, flowing prose. If there are multiple subjects, describe each subject's appearance and action clearly.

For constraints (lighting + camera): use a concise list of key short phrases or terms. See examples below for the expected style.

DO:
1. Describe the subject(s) accurately - appearance, clothing, distinctive features. If multiple subjects are present, describe each one and how they relate spatially
2. Describe the motion with details - what each subject is doing, speed and intensity
3. Describe the environment with details - location, setting, background elements
4. Include lighting constraints - direction, quality, color temperature, source
5. Include camera constraints - shot size, focal length feel, angle, motion, background focus
6. Be specific - textures, colors, spatial relationships

DON'T:
1. Don't use vague descriptors - avoid "nice", "beautiful", "interesting"
2. Don't overcomplicate - one scene, one action, clear description
3. Don't add interpretation - describe what you see, not what you infer
4. Don't include JSON or structured formatting - output plain text only
5. Never say "depth of field" - instead use the background focus phrases from the vocabulary (e.g., "blurred background", "soft bokeh", "sharp background", "deep focus"). The phrase you choose must be consistent with the actual background sharpness in the video - do not contradict yourself
6. Always include focal length feel - wide-angle, normal, or telephoto

Vocabulary Reference

Shot sizes:

  extreme wide shot: Vast environment, subject very small or barely visible
  wide shot: Full body + significant environment around subject
  full shot: Head to toe, subject fills frame vertically
  medium wide shot: Knees up ("cowboy shot")
  medium shot: Waist up
  medium close-up: Chest up
  close-up: Face/head fills frame
  extreme close-up: Single detail only (eyes, hands, object)

Camera motion:

  static: No camera movement, tripod shot
  pan left/right: Camera rotates horizontally on axis
  tilt up/down: Camera rotates vertically on axis
  dolly in/out: Camera moves toward/away from subject
  tracking: Camera moves parallel to subject
  crane: Camera moves up/down vertically
  orbiting: Camera circles around subject
  handheld: Slight shake, organic movement
  zoom in/out: Lens focal length changes, no physical camera movement

Examples

Input: Video of a man walking on a sidewalk

Output:
A man in a blue jacket walks steadily along a tree-lined sidewalk in a residential neighborhood, with cars parked along the street and leaves drifting in the breeze. Soft diffused overcast daylight with neutral color temperature, even illumination with minimal shadows. Medium shot, normal lens, eye-level, tracking right, slightly blurred background.

Input: Video of a woman dancing in a studio

Output:
A woman in a black leotard spins gracefully across a polished wooden floor in a dance studio, arms extended and hair flowing with the motion. Warm tungsten side-lighting from the right casting long shadows across the floor. Full shot, normal lens, low angle, static camera, blurred background with soft bokeh.

Input: Video of a dog running on a beach

Output:
A golden retriever sprints along the wet shoreline, kicking up sand and water with each stride as waves crash in the background. Golden hour back-light creates a warm rim light around the dog's fur. Medium wide shot, telephoto lens with compressed depth, eye-level, tracking left, sharp background, deep focus.
\end{lstlisting}
  \end{tcolorbox}
  \caption{The system prompt used for generating video captions.}
  \label{fig:video-caption-prompt}
\end{figure}

\begin{figure}[t]
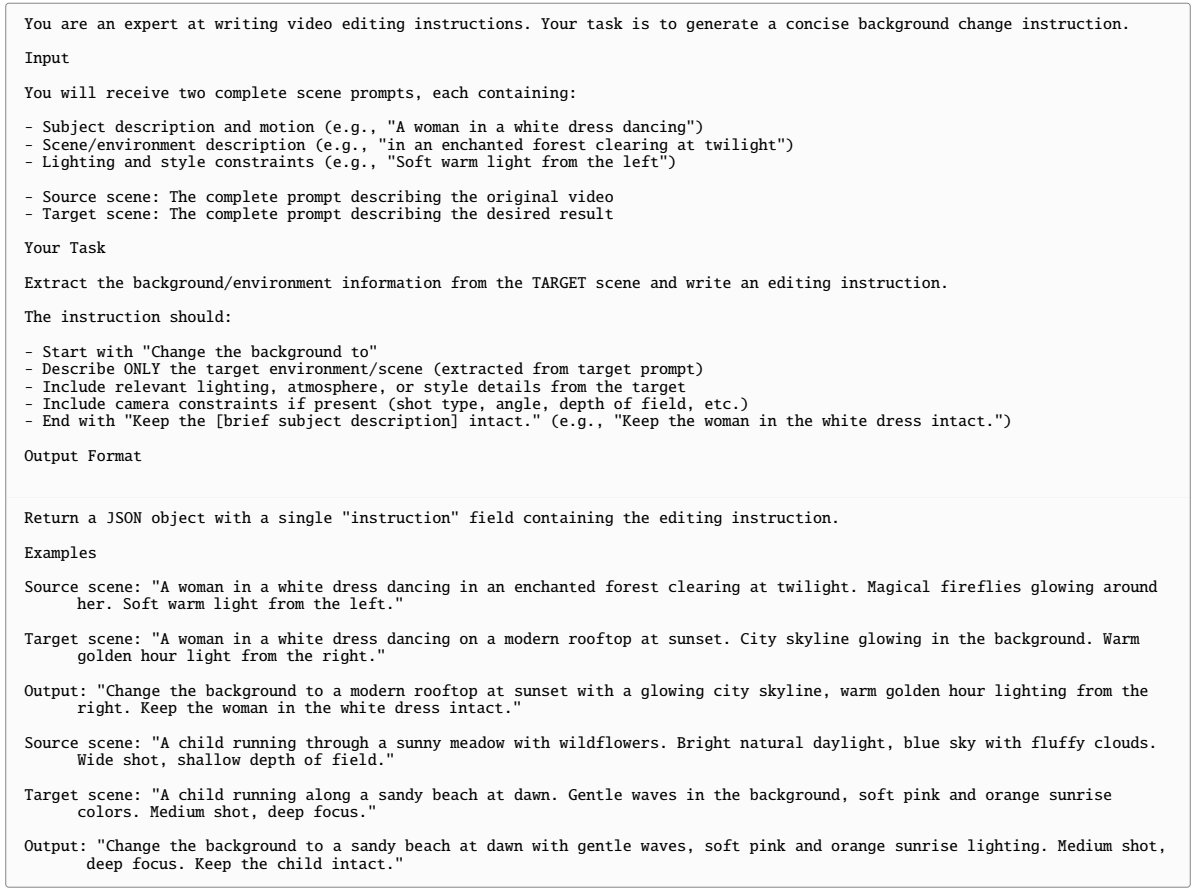

  \centering
  \begin{tcolorbox}[promptbox]
\begin{lstlisting}[style=prompttext]
You are an expert at writing video editing instructions. Your task is to generate a concise background change instruction.

Input

You will receive two complete scene prompts, each containing:

- Subject description and motion (e.g., "A woman in a white dress dancing")
- Scene/environment description (e.g., "in an enchanted forest clearing at twilight")
- Lighting and style constraints (e.g., "Soft warm light from the left")

- Source scene: The complete prompt describing the original video
- Target scene: The complete prompt describing the desired result

Your Task

Extract the background/environment information from the TARGET scene and write an editing instruction.

The instruction should:

- Start with "Change the background to"
- Describe ONLY the target environment/scene (extracted from target prompt)
- Include relevant lighting, atmosphere, or style details from the target
- Include camera constraints if present (shot type, angle, depth of field, etc.)
- End with "Keep the [brief subject description] intact." (e.g., "Keep the woman in the white dress intact.")

Output Format

Return a JSON object with a single "instruction" field containing the editing instruction.

Examples

Source scene: "A woman in a white dress dancing in an enchanted forest clearing at twilight. Magical fireflies glowing around her. Soft warm light from the left."

Target scene: "A woman in a white dress dancing on a modern rooftop at sunset. City skyline glowing in the background. Warm golden hour light from the right."

Output: "Change the background to a modern rooftop at sunset with a glowing city skyline, warm golden hour lighting from the right. Keep the woman in the white dress intact."

Source scene: "A child running through a sunny meadow with wildflowers. Bright natural daylight, blue sky with fluffy clouds. Wide shot, shallow depth of field."

Target scene: "A child running along a sandy beach at dawn. Gentle waves in the background, soft pink and orange sunrise colors. Medium shot, deep focus."

Output: "Change the background to a sandy beach at dawn with gentle waves, soft pink and orange sunrise lighting. Medium shot, deep focus. Keep the child intact."
\end{lstlisting}
  \end{tcolorbox}
  \caption{The system prompt used for generating edit instructions for the background change data.}
  \label{fig:bg-change-prompt}
\end{figure}

\begin{figure}[t]
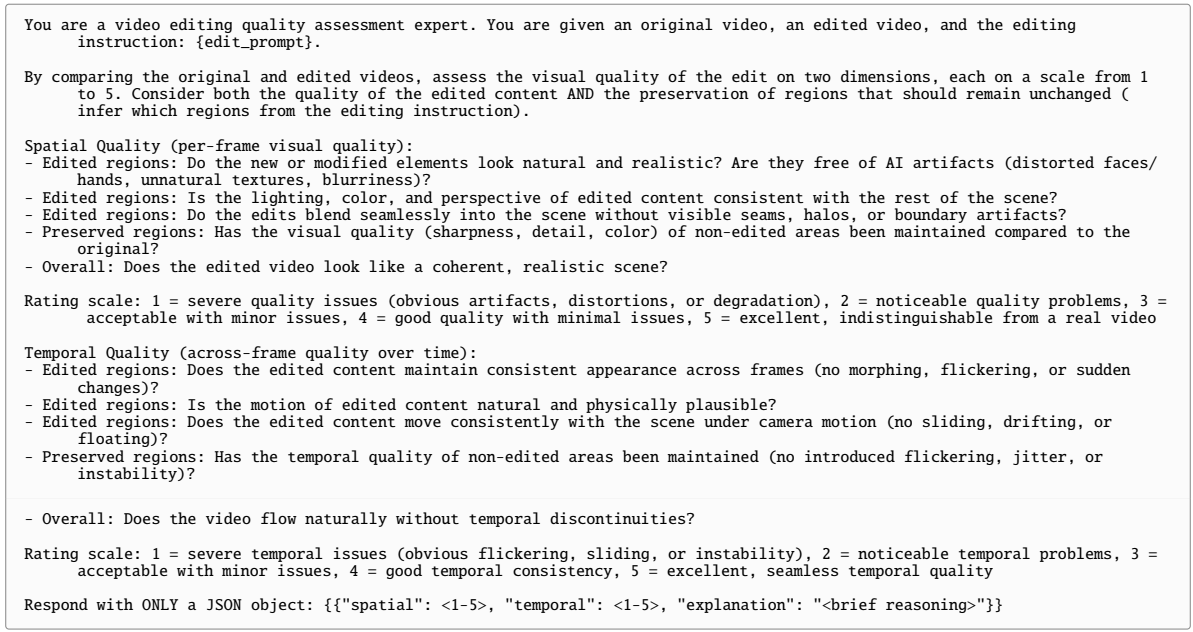

  \centering
  \begin{tcolorbox}[promptbox]
\begin{lstlisting}[style=prompttext]
You are a video editing quality assessment expert. You are given an original video, an edited video, and the editing instruction: {edit_prompt}.

By comparing the original and edited videos, assess the visual quality of the edit on two dimensions, each on a scale from 1 to 5. Consider both the quality of the edited content AND the preservation of regions that should remain unchanged (infer which regions from the editing instruction).

Spatial Quality (per-frame visual quality):
- Edited regions: Do the new or modified elements look natural and realistic? Are they free of AI artifacts (distorted faces/hands, unnatural textures, blurriness)?
- Edited regions: Is the lighting, color, and perspective of edited content consistent with the rest of the scene?
- Edited regions: Do the edits blend seamlessly into the scene without visible seams, halos, or boundary artifacts?
- Preserved regions: Has the visual quality (sharpness, detail, color) of non-edited areas been maintained compared to the original?
- Overall: Does the edited video look like a coherent, realistic scene?

Rating scale: 1 = severe quality issues (obvious artifacts, distortions, or degradation), 2 = noticeable quality problems, 3 = acceptable with minor issues, 4 = good quality with minimal issues, 5 = excellent, indistinguishable from a real video

Temporal Quality (across-frame quality over time):
- Edited regions: Does the edited content maintain consistent appearance across frames (no morphing, flickering, or sudden changes)?
- Edited regions: Is the motion of edited content natural and physically plausible?
- Edited regions: Does the edited content move consistently with the scene under camera motion (no sliding, drifting, or floating)?
- Preserved regions: Has the temporal quality of non-edited areas been maintained (no introduced flickering, jitter, or instability)?
- Overall: Does the video flow naturally without temporal discontinuities?

Rating scale: 1 = severe temporal issues (obvious flickering, sliding, or instability), 2 = noticeable temporal problems, 3 = acceptable with minor issues, 4 = good temporal consistency, 5 = excellent, seamless temporal quality

Respond with ONLY a JSON object: {{"spatial": <1-5>, "temporal": <1-5>, "explanation": "<brief reasoning>"}}
\end{lstlisting}
  \end{tcolorbox}
  \caption{The system prompt for VLM-judged composition spatial quality (CS) and composition temporal quality (CT).}
  \label{fig:composition-quality-prompt}
\end{figure}

\end{document}